\documentclass[10pt,twocolumn,letterpaper]{article}

\usepackage[pagenumbers]{cvpr} 

\usepackage{float}    
\usepackage{caption}
\usepackage{amsmath}
\usepackage{amssymb}
\usepackage{bm}
\usepackage{booktabs}
\usepackage{multirow}
\usepackage{makecell}
\usepackage{xspace}
\usepackage{colortbl}
\usepackage{placeins}  

\newcommand{\hp}{{\textsc{\small{{HetPoison}}}}\xspace}
\newcommand{\eps}{\varepsilon}
\newcommand{\fatt}{f_{\text{att}}}
\newcommand{\fego}{f_{\text{ego}}}
\newcommand{\Lrem}{\mathcal{L}_{\text{rem}}}
\newcommand{\pmstd}[1]{{\scriptsize$\pm$#1}}
\definecolor{hpcol}{RGB}{252,233,230}
\newcommand{\hpcell}{\cellcolor{hpcol}}
\newcommand{\best}[1]{\textbf{#1}}
\definecolor{tkcol}{RGB}{234,241,250}
\newcommand{\takeaway}[1]{\par\vspace{3pt}\noindent\colorbox{tkcol}{\parbox{\dimexpr\columnwidth-2\fboxsep\relax}{\small\textbf{Takeaway:} #1}}\par\vspace{3pt}}

\DeclareMathOperator*{\argmin}{arg\,min}

\definecolor{cvprblue}{rgb}{0.21,0.49,0.74}
\usepackage[pagebackref,breaklinks,colorlinks,allcolors=cvprblue]{hyperref}
\usepackage[capitalize]{cleveref}
\crefname{section}{Sec.}{Secs.}
\Crefname{section}{Section}{Sections}
\crefname{table}{Tab.}{Tabs.}
\Crefname{table}{Table}{Tables}

\def\paperID{17097} 
\def\confName{CVPR}
\def\confYear{2027}

\title{From Per-Frame Search to Practical Attacks: Investigating Adversarial Robustness of Multi-Agent Heterogeneous Cooperative Perception}

\author{
    Chenyi Wang, Yutong Liu, Qingzhao Zhang, Ming F. Li\\
    University of Arizona\\
    {\small\texttt{\{chenyiw, yutongl, qzzhang, lim\}@arizona.edu}}
}

\begin{document}
\maketitle

\begin{abstract}
Heterogeneous cooperative perception (CP) lets connected agents with different sensors and encoders share bird's-eye-view (BEV) features and align them through learned translation modules before fusion and detection.
A malicious collaborator can perturb its broadcast to erase real objects from a neighbor's view, yet prior CP attacks assume homogeneous agents and rely on per-frame gradient descent that needs test-time labels, online detector queries, and slow backward passes.
Porting these homogeneous attacks to heterogeneous CP exposes two further architectural barriers.
Translation modules often block direct gradients, and heterogeneous fusion stacks use per-location fusion and detection heads that leave the standard removal-loss gradient zero on over $99\%$ of the attacker's message, so per-frame updates reach only the sparse anchor footprint while neighboring cells still detect the object.
We present \hp, a practical feature attack that trains a convolutional generator once, offline, on a frozen surrogate stack to map channel-observable features to a bounded perturbation in one forward pass, requiring no labels, detector queries, or gradients online.
Because the generator parameterizes perturbations across the spatial neighborhood rather than per coordinate, \hp overcomes sparse anchor gradients while amortizing search across frames into a $\sim$2\,ms forward pass.
On OPV2V-H, \hp lowers STAMP detection average precision from $0.899$ to $0.063$, outperforms 500-step Adam-PGD on HEAL, STAMP, and CodeFilling at up to $14{,}000\times$ lower online cost, and remains effective on real-world V2X-Real fleets and under cross-family transfer, communication delay, quantization, and unseen collaborator encoders.
\end{abstract}

\section{Introduction}
\label{sec:intro}

\begin{figure*}[t]
\centering
\includegraphics[width=\textwidth]{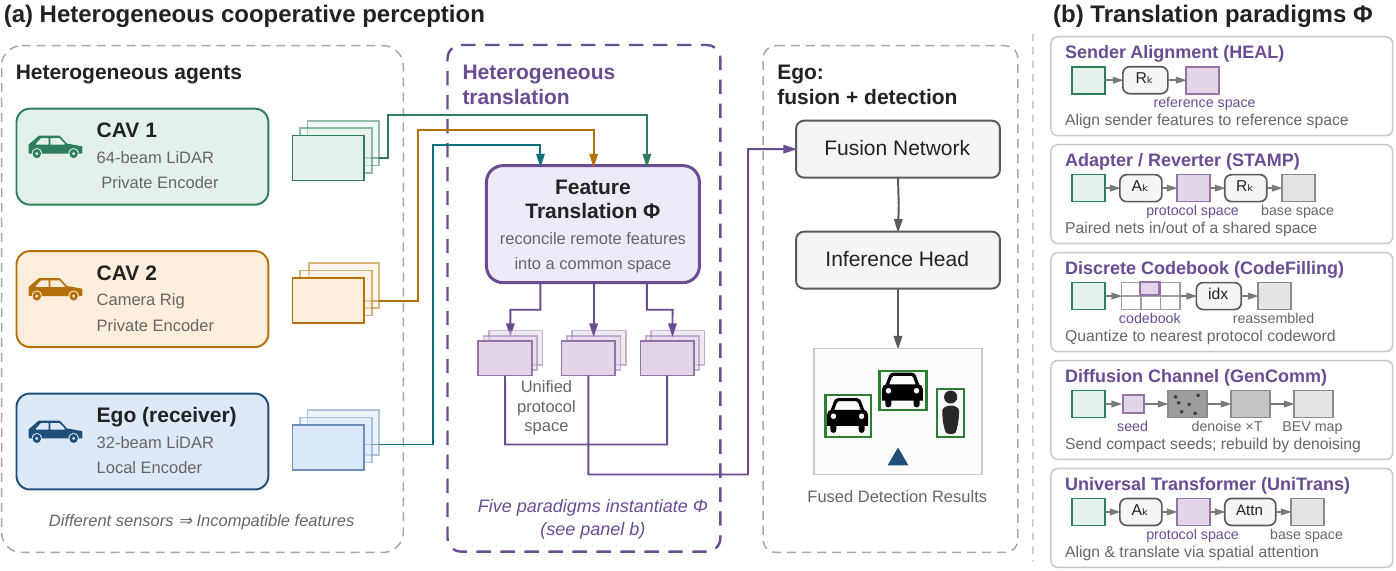}
\vspace{-5mm}
\caption{\textbf{Heterogeneous cooperative perception and its attack surface.} (a) Agents with different sensors and encoders broadcast BEV features; a translation module $\Phi$ maps them into the ego's space before fusion and detection. A malicious collaborator controls only its own broadcast. (b) Translation paradigms we evaluate. Sensors in (a) are illustrative; evaluated pairs are in \cref{sec:protocol}. HEAL aligns encoders at the sender, so its receiver-side $\Phi$ is the identity.}
\label{fig:hcp}
\vspace{-2mm}
\end{figure*}

A single agent system sees only what its own sensors reach, where a blocking object or a building corner can hide a crossing pedestrian. 
Cooperative perception (CP) removes such blind spots by letting networked agents, such as connected vehicles and roadside units, share intermediate bird's-eye-view (BEV) features over wireless links, which the receiver (the \emph{ego}) fuses before detection~\cite{opv2v2022,v2xvit2022,where2comm2022}.
Real-world fleets are heterogeneous, since agents carry different LiDARs and cameras and run proprietary encoders. 
Heterogeneous CP therefore introduces a learned translation module to map remote features into a space the ego can fuse via sender-side encoder alignment~\cite{lu2024heal,shao2024hetecooper,wang2026gtspace}, adapter and reverter pairs~\cite{stamp2025,luo2024plug,xia2025one,li2026one,shao2026negocollab}, discrete codebooks~\cite{codefilling2025,codealign2026}, or diffusion decoders~\cite{gencomm2025,CoDifFu} (\cref{fig:hcp}).

Feature sharing also opens a safety-critical attack surface, since a single compromised collaborator can broadcast perturbed features that inflate latency~\cite{cpfreezer2025} or make the ego miss real objects even when other agents see them clearly~\cite{tu2021advcomm,datafab2024,sombra2025,pretendbenign2025,mvig2026}.
However, existing CP attacks are designed for homogeneous systems where all agents share an identical sensor and encoder setup~\cite{tu2021advcomm,datafab2024,sombra2025,cpfreezer2025,pretendbenign2025,mvig2026}, and they face severe practicality challenges in deployment.
Because they solve a fresh projected gradient descent (PGD~\cite{madry2018pgd}) problem for every frame, they require real-time ground-truth labels, online backward passes through the victim's detector, and iterative computation that exceed the 100\,ms CP deadline~\cite{cpfreezer2025}.

\begin{figure*}[t]
\centering
\includegraphics[width=\textwidth]{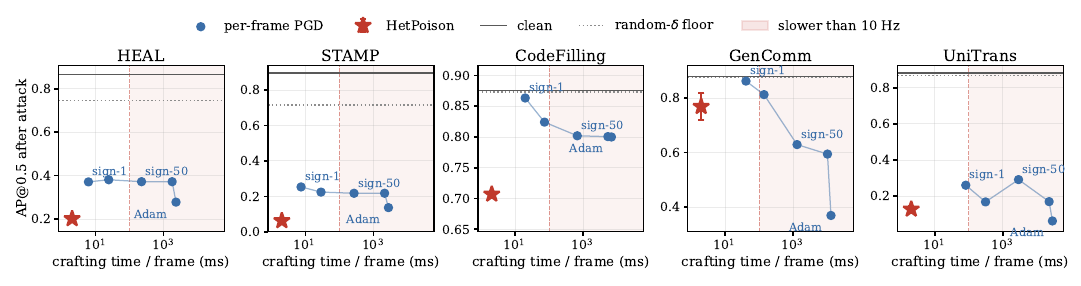}
\vspace{-6mm}
\caption{\textbf{Attack strength vs.\ crafting cost} on OPV2V-H at $\eps{=}1.0$ (\cref{sec:protocol}), measured per frame on an NVIDIA Jetson AGX Thor (\cref{tab:s-latency}). Blue: per-frame PGD with victim labels and heads (sign-PGD with 1, 5, 50, 400 steps; Adam-PGD with $5{\times}100$ steps). Red: \hp, one generator forward pass with no labels or queries online (mean $\pm$ std over 3 seeds; lower AP is stronger). The shaded region exceeds one 10\,Hz CP interval (100\,ms)~\cite{cpfreezer2025}. At $\eps{=}1.0$, \hp sits below the entire PGD frontier on HEAL, STAMP, and CodeFilling at $2.0$\,ms/frame (\cref{sec:why}); on GenComm and UniTrans, per-frame search overtakes \hp only after exceeding the 100\,ms deadline by $13$--$286\times$ (\cref{sec:why}).}
\label{fig:cost}
\vspace{-3mm}
\end{figure*}

Furthermore, porting these homogeneous per-frame attacks to heterogeneous CP exposes two additional architectural barriers that hinder coordinate-wise gradient search.
First, translation modules compress, discretize, or randomize incoming features before fusion, obstructing gradient backpropagation through discrete codebooks such as CodeFilling~\cite{codefilling2025}.
Second, per-frame optimization suffers from extreme gradient sparsity on heterogeneous fusion stacks.
On HEAL~\cite{lu2024heal}, STAMP~\cite{stamp2025}, and CodeFilling~\cite{codefilling2025}, whose translation modules are followed by per-location attention fusion and point-wise detection heads, the standard removal loss acts only on positive ground-truth anchors. 
The gradient with respect to the attacker's message is zero on over $99\%$ of its coordinates (\cref{sec:why}).
Per-frame PGD updates only that sub-percent anchor footprint, leaving adjacent cells unaffected, so they still produce detection boxes.

We overcome both practical and architectural barriers with \hp, an \emph{amortized} feature attack on heterogeneous CP (\cref{fig:method}).
Rather than optimizing a perturbation via gradient search online for each frame, \hp trains a convolutional generator once, offline, against a frozen surrogate stack to map channel-observable features to a bounded perturbation in one forward pass.
At attack time, it runs in $\sim$2\,ms on an embedded Jetson AGX Thor~\cite{nvidia_jetson_thor} without labels, victim queries, or gradients (\cref{fig:cost}).
Unlike coordinate-wise PGD, which leaves off-support coordinates at their initialization, \hp parameterizes the perturbation through a convolutional network whose receptive field couples positive anchors with their spatial neighborhood.
Writing the full message at once suppresses surrounding anchor predictions and traverses discrete codebooks where per-frame search saturates. To our knowledge, \hp is the first online data-fabrication attack against heterogeneous CP, and it tackles the architectural and deployability challenges together.

Evaluated under a unified protocol with matched perturbation budget, objective, and forward path across five heterogeneous paradigms on OPV2V-H~\cite{opv2v2022,lu2024heal} and real-world V2X-Real~\cite{v2xreal2024} benchmarks, \hp outperforms 500-gradient Adam-PGD on HEAL ($0.200$ vs.\ $0.277$ AP@0.5) and STAMP ($0.063$ vs.\ $0.136$) at $1200$--$14{,}000\times$ lower runtime cost, while beating every per-frame attack from zero or random starts on CodeFilling ($0.709$ vs.\ $0.800$).
Controlled ablations confirm that spatial convolutional parameterization recovers off-support neighborhood coverage even on a single frame, while cross-frame amortization eliminates online labels and delivers a four-order-of-magnitude speedup (\cref{sec:why}). 
On V2X-Real LiDAR--LiDAR and LiDAR--camera fleets, \hp reproduces these gains at $\sim$$3000\times$ lower crafting cost and remains effective across continuous cross-family transfer, communication delay, quantization, and unseen encoders. 
Our contributions are:
\begin{itemize}\itemsep0.5pt
\item We investigate the adversarial robustness of heterogeneous CP, tracing per-frame PGD saturation to step-size miscalibration and the sparse ($<1\%$) \textbf{gradient footprint} of per-location fusion and heads (\cref{sec:why}).
\item We propose \textbf{\hp}, a practical offline-trained feature attack on heterogeneous CP with single-pass real-time generation that needs no labels, queries, or gradients online (\cref{sec:method}), running in $\sim$2\,ms/frame on a Jetson AGX Thor (\cref{sec:main}).
\item We establish a \textbf{matched-objective attack harness} and unified protocol that fix the perturbation budget, removal objective, and translation forward path across matched-compute and adaptive PGD baselines, multi-seed runs, and object-level metrics (\cref{sec:protocol}). Code is available at \url{https://github.com/WiSeR-Lab/HetPoison} for public review and validation.
\item We evaluate across \textbf{five translation paradigms} under both matched and black-box transfer threat postures on OPV2V-H and V2X-Real, demonstrating \hp's effectiveness, practicality, and scope boundaries under channel imperfections (\cref{sec:results}).
\end{itemize}

\section{Background and Related Work}
\label{sec:related}

\textbf{Heterogeneous cooperative perception.}
Intermediate-fusion CP shares compact BEV features across networked agents such as connected vehicles and road-side units so the receiver can fuse multi-view context before detection~\cite{opv2v2022,v2xvit2022,where2comm2022,v2vam2022,coalign2023}.
In practice, such agents usually differ in sensor modalities (LiDARs with varying beam counts, multi-view cameras) and run proprietary backbones whose feature spaces differ in channel semantics, spatial resolution, and activation scale. 
Direct feature fusion across such agents fails without explicit cross-agent alignment.
Heterogeneous CP therefore introduces a learned translation module $\Phi$ that maps incoming collaborator features into a common representation space before ego-side fusion:
HEAL~\cite{lu2024heal,shao2024hetecooper,wang2026gtspace} aligns new agent types to a shared pyramid space by training only their sender-side encoders against a frozen base detector;
STAMP and UniTrans~\cite{stamp2025,li2026one,xia2025one,luo2024plug,shao2026negocollab} translate features across arbitrary agent pairs via sender-side adapters and receiver-side reverters or universal transformer modules around a shared protocol space; CodeFilling~\cite{codefilling2025,codealign2026} quantizes continuous BEV features to a shared discrete codebook and transmits compact code indices; and GenComm~\cite{gencomm2025,CoDifFu} transmits a two-channel spatial prior from which the receiver regenerates full BEV features via a conditional diffusion decoder.
We evaluate all these paradigms, spanning continuous, discrete, stochastic, and attention translation.

\vspace{2pt}\noindent\textbf{Attacks on cooperative perception.}
Tu \etal~\cite{tu2021advcomm} and Zhang \etal~\cite{datafab2024} showed that shared CP features are vulnerable to data fabrication through per-frame PGD perturbations.
Subsequent work exploits attention weights~\cite{sombra2025}, inflates perception latency~\cite{cpfreezer2025}, hides perturbations from consistency checks~\cite{pretendbenign2025}, or adapts attack placement to the victim's viewpoint~\cite{mvig2026}. 
However, these attacks target homogeneous agents and perform gradient optimization per frame, assuming online white-box access and test-time labels.
Even when prior homogeneous attacks reuse perturbations across adjacent frames for real-time execution~\cite{datafab2024}, optimize attention/viewpoint placement~\cite{sombra2025,mvig2026}, or evade consistency detectors~\cite{pretendbenign2025}, 
their online perturbation updates still rely on coordinate-wise PGD with scalar step sizes and do not address the $<1\%$ gradient footprint or discrete/stochastic bottlenecks of heterogeneous translation modules.
\hp is orthogonal to placement and stealth objectives, replacing per-frame coordinate search with an offline-trained spatial generator for real-time online feature fabrication in heterogeneous CP. 

\vspace{2pt}\noindent\textbf{Generative and amortized attacks.}
Training a network to output adversarial perturbations has shown feasibility for tasks such as image classification and segmentation~\cite{gap2018, advgan2018}, outperforming input-independent universal adversarial perturbations~\cite{uap2017}.
Later work demonstrates generative perturbations' transferability across models and domains~\cite{naseer2019cda} and applicability to single-agent LiDAR detectors~\cite{lidaradvgan2025,advganlidarobst2023}.
These are instances of amortized optimization~\cite{amos2023amortized}, where a model predicts solutions across a family of problems. 
We bring this principle to multi-agent feature sharing and isolate the roles of convolutional parameterization and amortization across heterogeneous translators.

\section{Problem Setup}
\label{sec:setup}

\textbf{System model.}
As shown in \cref{fig:hcp}, each agent $k$ encodes its sensor data into a BEV feature $f_k$ with its own backbone and broadcasts it (\eg through C-V2X link~\cite{CV2X_ETSI_EN_303_613}). 
The ego agent projects received features via a translation module $\Phi$ into a unified protocol space, fuses with its own $\fego$, and runs its detection heads $h$ to produce perception results:
\begin{equation}
\hat{y} = h\big(\mathrm{fuse}(\fego, \{\Phi(f_k)\}_{k\neq\text{ego}})\big).    
\end{equation}

\vspace{2pt}\noindent\textbf{Threat model.}
The attacker controls one collaborator, either as an authenticated participant with a compromised software stack or on its communication link, which message signing cannot prevent~\cite{datafab2024,v2x-misbehave}.
Instead of its clean feature $\fatt$, it transmits a perturbed message $\fatt+\delta$.
\emph{Knowledge and threat postures:} offline, the attacker trains against a differentiable surrogate CP stack on its own driving logs (\cref{sec:method}). We formulate two threat postures: \textbf{matched posture}, where the surrogate shares the victim's open-protocol family (establishing a worst-case bound with the same architectural knowledge as white-box PGD, but computed offline); and \textbf{transfer posture}, where the victim stack is completely unknown (\cref{sec:deploy}). Online, the attacker observes only channel broadcasts ($\fatt$, $\fego$, and $\{f_k\}$; relaxed to lagged broadcasts in \cref{sec:deploy}).
\emph{Constraints:} zero queries, zero gradients, and zero labels online; runtime execution must strictly satisfy the 100\,ms CP deadline~\cite{cpfreezer2025}.
\emph{Goal:} object removal, causing the ego to miss true objects~\cite{tu2021advcomm,sombra2025}.

\vspace{2pt}\noindent\textbf{Perturbation budget.}
All attacks share one elementwise relative budget, scaled by the ego's feature so that a single $\eps$ is meaningful across encoders with different activation scales:
\begin{equation}
|\delta_i| \le \eps_i = \eps\cdot\max\!\big(|f_{\text{ego},i}|,\; \tfrac{1}{d}\textstyle\sum_j |f_{\text{ego},j}|\big),
\label{eq:budget}
\end{equation}
where the mean over all $d$ coordinates keeps the ball non-degenerate where $\fego$ is near zero.
Scaling by the receiver's feature on the shared grid makes $\eps$ comparable across different perturbation optimization strategies. 

\vspace{2pt}\noindent\textbf{Removal objective.}
Given labels, the removal loss is the mean objectness over anchors $\mathcal{P}$ marked positive by the ground truth,
$\Lrem(\hat{y}) = \frac{1}{|\mathcal{P}|}\sum_{a\in\mathcal{P}} \sigma(\text{cls}_a)$.
Per-frame attacks minimize $\Lrem$ online on the victim's own heads; \hp minimizes it offline on surrogate heads.

\section{\hp}
\label{sec:method}

\begin{figure}[t]
\centering
\includegraphics[width=0.92\linewidth]{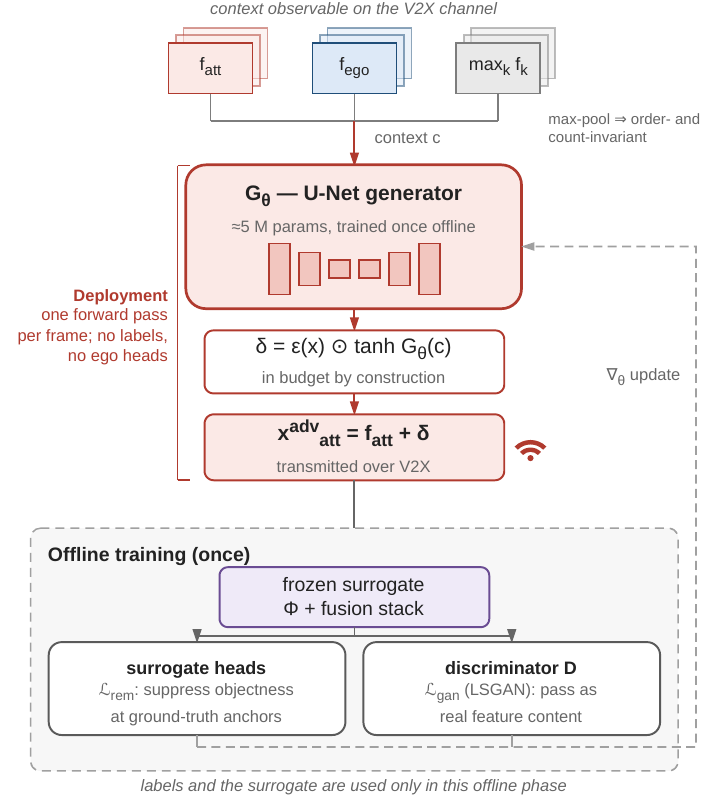}
\vspace{-3mm}
\caption{\textbf{\hp.} A U-Net $G_\theta$ maps the channel-observable context to a perturbation that lies in the budget by construction. It is trained once, offline, through a frozen surrogate CP stack; at attack time it runs only a single forward pass.}
\label{fig:method}
\vspace{-4mm}
\end{figure}

\hp turns the per-frame feature attack from expensive gradient-based search into a single-forward prediction (\cref{fig:method}).
From the shared features that an on-channel attacker observes at inference ($\fatt$, $\fego$, and other collaborators' broadcasts), a convolutional generator $G_\theta$ outputs a bounded perturbation $\delta$ added to $\fatt$ before transmission so the ego misses objects it would otherwise detect.
Trained once offline against a frozen surrogate stack, $G_\theta$ uses no labels, queries, or gradients online.

\vspace{2pt}\noindent\textbf{Amortized formulation.}
A per-frame attacker iteratively solves $\delta^\star(x) = \argmin_{|\delta|\le\eps(x)} \Lrem(\hat{y}(x,\fatt+\delta))$ via tens to hundreds of gradient steps through victim heads and labels.
\hp replaces this inner search with a learned map fitted over driving frames~\cite{amos2023amortized}:
\begin{equation}
\min_\theta\; \mathbb{E}_{x\sim\mathcal{D}_{\text{att}}}\Big[\Lrem\big(\tilde{y}(x,\fatt+\delta_\theta(x))\big)\Big] + \lambda\,\mathcal{L}_{\text{gan}},
\label{eq:amortized}
\end{equation}
where $\tilde{y}$ is the frozen surrogate output and $\mathcal{D}_{\text{att}}$ is the offline training split (never evaluation frames).
Because $\delta_\theta$ is parameterized by a convolutional network with spatially shared weights, it predicts a coherent perturbation field rather than per-coordinate updates (\cref{sec:why}).

\vspace{2pt}\noindent\textbf{Perturbation model and conditioning.}
$G_\theta$ conditions on the channel-observable context
$c = [\,\fatt \,\Vert\, \fego \,\Vert\, \max_k f_k\,]$,
concatenating the attacker's message, the ego's broadcast, and an elementwise max over remaining collaborators so one generator serves any number and order of agents.
The perturbation is
\begin{equation}
\delta_\theta(x) = \eps(x)\odot\tanh\!\big(G_\theta(c)\big),
\label{eq:delta}
\end{equation}
with $\eps(x)$ from \cref{eq:budget}, guaranteeing $|\delta_\theta(x)|\le\eps(x)$ without projection.
$G_\theta$ is a three-scale U-Net~\cite{ronneberger2015unet} with three residual blocks ($5.1$\,M parameters on $128$-channel features) that accepts any BEV resolution and writes across each object's spatial neighborhood via its receptive field (perturbing GenComm's two-channel diffusion seed directly).

\vspace{2pt}\noindent\textbf{Feature distribution regularization.}
Optimizing $\Lrem$ alone can push $\fatt+\delta$ off the clean feature manifold expected by translation and fusion.
We co-train a least-squares PatchGAN~\cite{lsgan2017,advgan2018} whose loss $\mathcal{L}_{\text{gan}}$ pulls local patches of $\fatt+\delta$ toward normal collaborator features.
This preserves the clean manifold (\cref{tab:s-gan-manifold}), sharpens victim attack (\cref{tab:s-ablation}), and crucially evades feature-space anomaly detection: against SOTA defense LUCIA~\cite{sombra2025}, removing $\mathcal{L}_{\text{gan}}$ allows its outlier filter to detect the perturbation, collapsing STAMP removal success from $0.325$ to $0.020$ (\cref{sec:s-discussion}).
All victims share one recipe ($1{,}000$ single-frame Adam updates, lr $2{\times}10^{-4}$, 3 seeds; \cref{sec:s-impl}).

\begin{table*}[t]
\centering
\small
\setlength{\tabcolsep}{2.6pt}
\caption{\textbf{Results on OPV2V-H with matched harness} (AP@0.5 $\downarrow$; lower is stronger). Per-frame attacks (s-$k$: sign-PGD-$k$; Adam: $5{\times}100$ steps) use current-frame labels and victim heads online; \hp uses an offline surrogate and neither labels nor gradients online. Bold / underline: strongest / weakest attack (excluding Floor). Supplement provides step sweeps, adaptive attacks, and AP@0.7 in \cref{tab:s-full,tab:s-adaptive,tab:s-ap07}. \hp 3-seed std is $\le0.007$ except UniTrans ($0.024$/$0.013$) and GenComm ($0.049$/$0.031$). Bottom: median time/frame on an NVIDIA Jetson AGX Thor (per-victim breakdown in \cref{tab:s-latency}).}
\label{tab:main}
\begin{tabular}{@{}l|c|cccccc||cccccc@{}}
\toprule
& & \multicolumn{6}{c||}{$\eps=1.0$} & \multicolumn{6}{c}{$\eps=1.5$} \\
Victim & Clean & Floor & s-1 & s-50$^\dagger$ & s-50 & Adam & \hp & Floor & s-1 & s-50$^\dagger$ & s-50 & Adam & \hp \\
\midrule
HEAL        & .866 & .744 & .374 & \underline{.822} & .373 & .277 & \hpcell\best{.200} & .400 & .279 & \underline{.754} & .293 & \best{.088} & \hpcell.130 \\
STAMP       & .899 & .716 & .255 & \underline{.727} & .217 & .136 & \hpcell\best{.063} & .342 & .189 & \underline{.537} & .158 & \best{.022} & \hpcell.032 \\
CodeFilling & .876 & .873 & \underline{.863} & .848 & .803 & .800 & \hpcell\best{.709} & .872 & \underline{.857} & .823 & .786 & .784 & \hpcell\best{.646} \\
GenComm     & .880 & .877 & \underline{.863} & .785 & .630 & \best{.369} & \hpcell.768 & .874 & \underline{.858} & .619 & .395 & \best{.076} & \hpcell.655 \\
UniTrans    & .882 & .866 & .263 & \underline{.299} & .288 & \best{.062} & \hpcell.128 & .757 & .038 & \underline{.289} & .201 & \best{.018} & \hpcell.078 \\
\midrule
V2VAM (homo.) & .957 & .946 & .417 & \underline{.630} & .081 & \best{.030} & \hpcell.043 & .929 & .208 & \underline{.447} & .040 & \best{.010} & \hpcell.017 \\
\midrule
ms / frame & -- & -- & 6--81 & \multicolumn{2}{c}{220--2900} & \underline{2300--28600} & \textbf{2.0} & \multicolumn{6}{c}{\footnotesize (crafting time does not depend on $\eps$)} \\
\bottomrule
\end{tabular}\\[1pt]
{\footnotesize $^\dagger$\,Scalar-step sign-PGD-50 ($2.5\,\overline{\eps}/50$), the optimization strategy adopted by prior homogeneous CP attacks~\cite{tu2021advcomm,datafab2024,sombra2025, pretendbenign2025, mvig2026}.}
\vspace{-4mm}
\end{table*}

\vspace{2pt}\noindent\textbf{Training through heterogeneous translators.}
Gradients flow from surrogate heads through frozen fusion and translation into $G_\theta$.
Non-differentiable and stochastic translators are traversed as the receiver runs them: CodeFilling uses a straight-through estimator, and GenComm backpropagates through diffusion sampling with fresh noise draws per pass (\cref{sec:protocol}).

\vspace{2pt}\noindent\textbf{Threat postures.}
In the \emph{matched posture}, the offline surrogate shares the victim's protocol family (matching PGD's architectural access offline for worst-case analysis); in the \emph{transfer posture}, the victim family is unknown (\cref{sec:deploy}). Both operate strictly query- and label-free online.

\section{Evaluation}
\label{sec:results}

\subsection{Experimental Setup}
\label{sec:protocol}

\textbf{Datasets.}
We evaluate on two multi-agent heterogeneous CP benchmarks:
\emph{OPV2V-H}~\cite{opv2v2022,lu2024heal}, which extends CARLA-simulated OPV2V~\cite{opv2v2022} with heterogeneous sensors across urban and highway scenes, and
\emph{V2X-Real}~\cite{v2xreal2024}, a real-world dataset from two connected vehicles and two roadside units with heterogeneous LiDARs and cameras.

\vspace{2pt}\noindent\textbf{Victim models.}
Across both datasets, we evaluate five heterogeneous CP families: \emph{HEAL}~\cite{lu2024heal} (sender pyramid alignment), \emph{STAMP}~\cite{stamp2025} (adapter--reverter networks), \emph{CodeFilling}~\cite{codefilling2025} (discrete codebook quantization), \emph{GenComm}~\cite{gencomm2025} (conditional diffusion decoding), and \emph{UniTrans}~\cite{li2026one} (transformer universal translation).
\textit{On OPV2V-H}: HEAL, CodeFilling, GenComm, and UniTrans pair a 64-beam PointPillars~\cite{lang2019pointpillars} LiDAR ego with an LSS~\cite{lss} camera collaborator (m1m2), STAMP pairs two heterogeneous LiDARs (m0m1), and \emph{V2VAM}~\cite{v2vam2022}, \emph{CoAlign}~\cite{coalign2023}, and \emph{AttFuse}~\cite{opv2v2022} serve as homogeneous LiDAR controls (\cref{tab:s-homo}).
\textit{On V2X-Real}, all five models are evaluated under \emph{LiDAR--LiDAR (LL)} (vehicle PointPillars + roadside SECOND) and \emph{LiDAR--camera (LC)} (PointPillars + LSS camera; \cref{sec:s-v2xreal}).

\vspace{2pt}\noindent\textbf{Matched-objective attack harness.}
To separate architectural robustness from gradient obfuscation, our \emph{matched-objective attack harness} fixes per frame and victim (\cref{sec:s-protocol}):
(1)~\textbf{The Ball} (perturbation budget), $|\delta|\le\eps\cdot\max(|f_{\text{ego}}|,\overline{|f_{\text{ego}}|})$ (\cref{eq:budget});
(2)~\textbf{The Objective}, the ground-truth removal loss $\Lrem$ (\cref{sec:setup}) on the ego's heads; and
(3)~\textbf{The Forward Path}, the victim's true translation $\Phi$, spatial alignment, and fusion stack, attributing any gap in attack success solely to the optimizer.

\vspace{2pt}\noindent\textbf{Baselines.}
Per-frame baselines assume an online white-box posture with test-time labels and detector backpropagation (\cref{sec:s-protocol}); \hp is strictly black-box and label-free online, precomputing perturbations offline.
\emph{Random-$\delta$ floor} draws uniform noise in the ball of \cref{eq:budget} as a non-optimization reference.
\emph{sign-PGD-$k$}~\cite{madry2018pgd} ($k\in\{1,5,50,400\}$) represents the standard sign-gradient attack from prior homogeneous CP work~\cite{tu2021advcomm,datafab2024,sombra2025,pretendbenign2025,mvig2026}.
Whereas prior work uses a uniform \emph{scalar step} $2.5\,\overline{\eps}/k$ (\emph{s-50$^\dagger$} in \cref{tab:main}), which remains effective when gradients are dense (\eg, V2VAM and UniTrans), it severely under-steps high-activation foreground coordinates in heterogeneous CP and leaves attack on HEAL ($0.822$) and STAMP ($0.727$) weaker than random noise (\cref{sec:why}); we therefore equip sign-PGD-$k$ with a calibrated \emph{elementwise step} $2.5\,\eps_i/k$ from zero so every coordinate can reach $\pm\eps_i$.
\emph{Adam-PGD} is our strongest coordinate-wise baseline, combining adaptive moments~\cite{kingma2015adam} (step $2.5\,\eps_i/100$) to normalize channel gradient scales with $5{\times}100$ restarts (one zero and four random starts, 500 gradients/frame) to escape local minima and perturb zero-gradient coordinates.
Adaptive white-box attacks (\emph{EOT-8}~\cite{athalye2018eot} and \emph{BPDA}~\cite{athalye2018obfuscated}) for stochastic and discrete translators are evaluated in Supplement \cref{sec:s-opv2v}.

\vspace{2pt}\noindent\textbf{Metrics and variance.}
We report 3D AP@0.5 $\downarrow$ (AP@0.7 in \cref{tab:s-ap07}) over all objects on OPV2V-H and vehicles on V2X-Real, object-level \emph{recall}, \emph{removal success (RS)} (overall and ego-only), \emph{false positives (FP)}/frame, and median per-frame \emph{crafting cost} on a Jetson AGX Thor (\cref{sec:s-protocol}).

\begin{figure*}[t]
\centering
\includegraphics[width=0.95\textwidth]{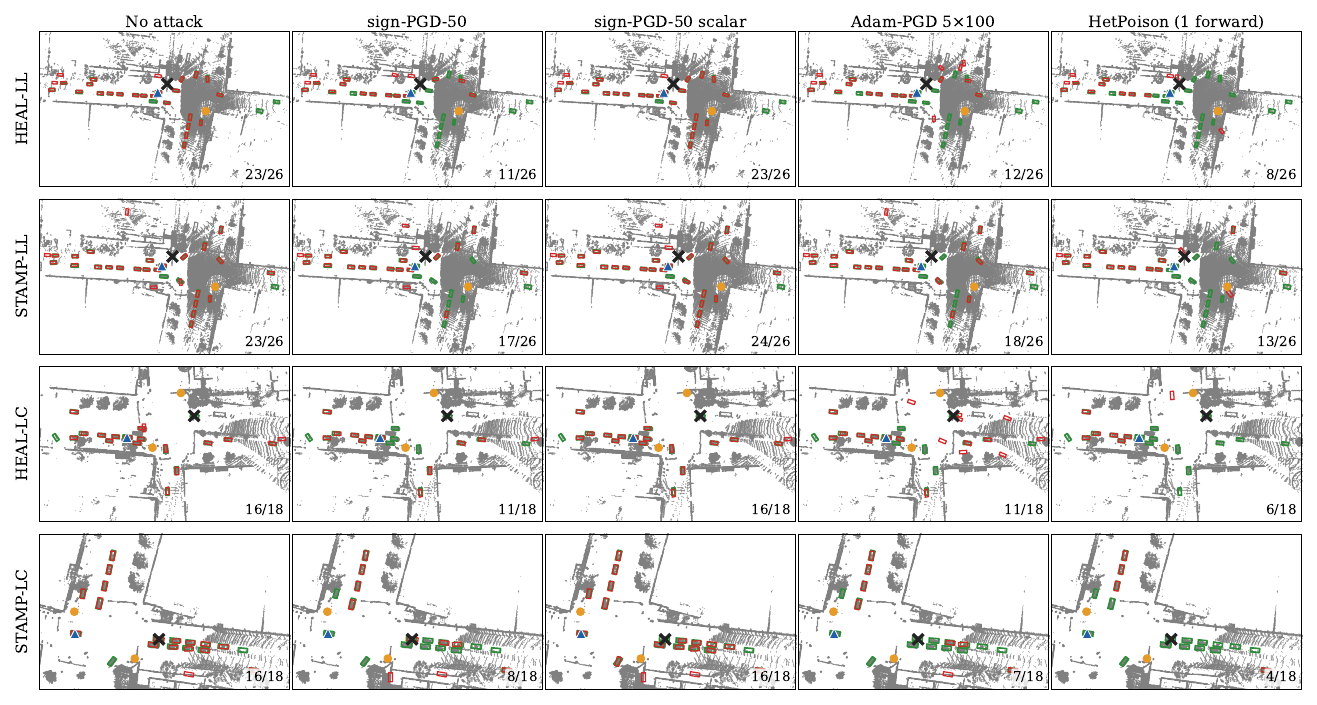}
\vspace{-3mm}
\caption{\textbf{Qualitative visualization on V2X-Real}. Green: ground truth; red: fused detections; {\color[RGB]{31,95,168}$\blacktriangle$} ego; $\times$ attacker; {\color[RGB]{230,154,40}$\bullet$} benign collaborator. Numbers: vehicles detected (IoU $\ge 0.5$) / ground truth. We provide OPV2V-H visualizations in Supplement \cref{fig:qual}.}
\label{fig:qual-v2x}
\vspace{-4mm}
\end{figure*}

\subsection{Main Results on OPV2V-H}
\label{sec:main}

\textbf{Continuous alignment and adapter--reverter translators.}
Across OPV2V-H (\cref{tab:main}, \cref{fig:cost}), attack outcomes divide by whether translation and fusion confine gradients to positive anchors or spread them spatially (\cref{sec:why}).
On HEAL and STAMP at $\eps{=}1.0$, \hp lowers AP from $0.866$ to $0.200$ and from $0.899$ to $0.063$, ahead of 500-step Adam-PGD by $0.07$--$0.08$ AP, with rankings unchanged at AP@0.7 (\cref{tab:s-ap07}).
The scalar-step s-50$^\dagger$ is weaker than random noise, elementwise sign-PGD saturates after one step (\cref{tab:s-full}), and only at $\eps{=}1.5$ does Adam-PGD edge ahead ($0.01$--$0.04$ AP).

\vspace{2pt}\noindent\textbf{Discrete codebook translation.}
On CodeFilling, all per-frame baselines, including adaptive EOT-8/BPDA (\cref{tab:s-adaptive}) and larger budgets $\eps\in\{2,3,5\}$, plateau at $0.78$--$0.87$ AP, a classic gradient-masking signature~\cite{athalye2018obfuscated,croce2020autoattack,carlini2019evaluating}.
\hp moves past this plateau ($0.709$ and $0.646$) by changing $73$--$89\%$ of code indices vs.\ $14$--$22\%$ for Adam-PGD (\cref{tab:s-flip}).
An unconstrained code-flip oracle reaches only $0.686$ (\cref{tab:s-oracle}), so \hp already attains ${\sim}89\%$ of the AP drop the discrete channel can carry.

\begin{table}[t]
\centering
\footnotesize
\setlength{\tabcolsep}{2.2pt}
\caption{\textbf{Object-level metrics} on OPV2V-H ($\eps{=}1.0$). Rec.: recall clean$\to$attacked (\%). RS: removal success (\%), overall and ego-only. FP: false positives/frame (clean in parentheses; \hp suppresses FP below clean).}
\label{tab:object}
\begin{tabular}{@{}ll|cc|c|c@{}}
\toprule
Victim & Attack & Rec. & RS & RS (ego only) & FP \\
\midrule
\multirow{3}{*}{HEAL} & sign-50 & 88$\to$43 & 52.0 & 22.4 & 4.2 (4.1) \\
 & Adam & 88$\to$40 & 54.8 & 23.7 & 9.4 (4.2) \\
 & \hp & \hpcell88$\to$22 & \hpcell\best{75.7} & \hpcell\best{38.9} & \hpcell2.5 (4.1) \\
\midrule
\multirow{3}{*}{STAMP} & sign-50 & 92$\to$29 & 68.9 & 29.9 & 3.7 (4.2) \\
 & Adam & 92$\to$28 & 70.3 & 31.8 & 9.7 (4.2) \\
 & \hp & \hpcell92$\to$9 & \hpcell\best{90.8} & \hpcell\best{50.5} & \hpcell2.1 (4.2) \\
\midrule
\multirow{3}{*}{\makecell[l]{Code-\\Filling}} & sign-50 & 90$\to$84 & 6.5 & 2.8 & 6.4 (6.3) \\
 & Adam & 90$\to$84 & 6.9 & 2.6 & 6.2 (6.3) \\
 & \hp & \hpcell90$\to$74 & \hpcell\best{18.2} & \hpcell\best{13.1} & \hpcell3.9 (6.3) \\
\midrule
\multirow{3}{*}{GenComm} & sign-50 & 90$\to$74 & 17.7 & 10.3 & 5.8 (3.4) \\
 & Adam & 90$\to$54 & \best{39.9} & \best{30.3} & 8.0 (3.4) \\
 & \hp & \hpcell90$\to$83 & \hpcell8.2 & \hpcell2.1 & \hpcell4.6 (3.4) \\
\midrule
\multirow{3}{*}{UniTrans} & sign-50 & 90$\to$45 & 50.4 & 56.3 & 11.2 (5.3) \\
 & Adam & 90$\to$20 & \best{77.9} & \best{63.9} & 31.7 (5.3) \\
 & \hp & \hpcell90$\to$22 & \hpcell75.2 & \hpcell32.3 & \hpcell55.1 (5.3) \\
\bottomrule
\end{tabular}
\vspace{-2mm}
\end{table}

\takeaway{Per-frame PGD substantially \emph{overestimates} the robustness of continuous and discrete translators. A single generator pass is the strongest evaluated method on HEAL, STAMP, and CodeFilling, and on CodeFilling it nearly reaches the channel's information ceiling.}

\vspace{2pt}\noindent\textbf{Diffusion and transformer translators and homogeneous controls.}
When a spatial network follows $\delta$ (GenComm's diffusion decoder, UniTrans's transformer, or V2VAM's backbone), gradients are dense (\cref{sec:why}), letting white-box search exploit per-frame labels.
On GenComm, Adam-PGD penetrates the diffusion channel, whereas \hp stays at $0.768$ ($\eps{=}1.0$) as single-pass generation cannot adapt to online diffusion noise (\cref{sec:s-opv2v}).
On UniTrans, \hp cuts AP from $0.882$ to $0.128$ in $2.0$\,ms without labels or gradients, against Adam-PGD's $0.062$ and s-1's $0.263$ at $81.3$\,ms with labels; at $\eps{=}1.5$ it removes more objects than either per-frame attack ($84.6\%$ vs.\ $77.1\%$; \cref{tab:s-object}).
On homogeneous V2VAM, \hp closely matches Adam-PGD within $0.013$ AP (\cref{tab:main}).

\vspace{2pt}\noindent\textbf{Computational cost and real-time feasibility.}
On a Jetson AGX Thor (\cref{tab:main,tab:s-latency}, \cref{fig:cost}), \hp executes in $2.0$\,ms/frame, well within the 100\,ms (10\,Hz) V2X deadline~\cite{cpfreezer2025}, while s-1 takes $6$--$81$\,ms and Adam-PGD $2.3$--$28.6$\,s ($1{,}200$--$14{,}000\times$ slower).
At $\eps{=}1.0$, per-frame search overtakes \hp on GenComm/UniTrans only after breaching the deadline by $13$--$286\times$ (\cref{fig:cost}); at $\eps{=}1.5$, UniTrans s-1 craft ($81.3$\,ms) plus downstream fusion ($41.9$\,ms) totals $123.2$\,ms, still violating the 100\,ms budget (\cref{tab:s-latency}).
Offline training ($6$--$16$\,min) amortizes in $70$--$390$ frames (\cref{sec:s-protocol}).

\vspace{2pt}\noindent\textbf{Object removal and false-positive suppression.}
At the object level (\cref{tab:object,tab:s-object}), \hp removes $76\%$ (HEAL) and $91\%$ (STAMP) of detected objects vs.\ $55\%$ and $70\%$ for Adam-PGD, while remarkably \emph{suppressing} false positives below clean driving ($2.5$ vs.\ $4.1$ clean on HEAL; $2.1$ vs.\ $4.2$ on STAMP) where Adam-PGD inflates them to $9.4$--$9.7$/frame.
Coordinate-wise PGD optimizes anchors independently, creating noisy high-frequency spikes that trigger phantom detections; \hp's convolutional field and $\mathcal{L}_{\text{gan}}$ silence boxes smoothly on the clean feature manifold.
On UniTrans ($\eps{=}1.5$), \hp removes more objects ($84.6\%$ vs.\ $77.1\%$) with $4\times$ fewer false positives; on CodeFilling, it doubles removed objects vs.\ PGD (\cref{tab:s-object}).
Crucially, \hp erases $39$--$51\%$ of objects visible \emph{only} to the ego's own LiDAR and $>75\%$ of jointly visible objects (\cref{fig:qual,tab:s-object}), actively corrupting local ego perception.

\begin{table}[t]
\centering
\footnotesize
\setlength{\tabcolsep}{1.5pt}
\caption{\textbf{Parameterization and amortization} (OPV2V-H, $\eps{=}1.0$, AP@0.5 $\downarrow$). \hp: seed 0 (3 seeds: $.201$, $.063$, $.706$, $.776$, $.131$); per-frame $G$: 3 seeds (500 Adam steps/frame). Support: non-zero $\nabla_{\fatt}\Lrem$ share at $\delta{=}0$ (\cref{tab:s-support}).}
\label{tab:amort}
\begin{tabular}{@{}l|ccccc@{}}
\toprule
& HEAL & STAMP & CodeF. & GenC. & UniTr. \\
\midrule
Gradient support & 0.66\% & 0.73\% & 0.71\% & 100\% & 100\% \\
\midrule
Adam-PGD                 & .271 & .133 & .801 & .369 & \best{.064} \\
From \hp's $\delta$\textsuperscript{$\dagger$} & .209 & .066 & .688 & .347 & .038 \\
Per-frame $G$            & .217 & .082 & .710 & \best{.350} & .072 \\
\hp (1 forward)          & \hpcell\best{.201} & \hpcell\best{.058} & \hpcell\best{.704} & \hpcell.738 & \hpcell.139 \\
\hp on the support only  & .326 & .154 & .791 & .738 & .139 \\
\bottomrule
\end{tabular}\\[1pt]
\parbox{\columnwidth}{\footnotesize $^\dagger$\,Oracle warm-start: 500-step Adam-PGD initialized at \hp's $\delta$ to test whether per-frame PGD's gap stems from off-support initialization rather than just step calibration.}
\vspace{-2mm}
\end{table}

\subsection{Mechanism and Ablation Analysis}
\label{sec:why}

\textbf{Step-size miscalibration and sparse gradient support.}
Two mechanisms cause standard per-frame PGD to overestimate heterogeneous CP robustness.
First, on HEAL, STAMP, and CodeFilling, $\nabla_{\fatt}\Lrem$ is non-zero on only $0.66$--$0.73\%$ of coordinates (\cref{tab:amort,tab:s-support}), because spatial warps, per-location attention, and $1{\times}1$ heads follow $\fatt$, restricting positive anchors to their own BEV cells. Conversely, when a spatial network follows $\delta$, gradients are dense (GenComm $100\%$, UniTrans $100\%$; \cref{tab:s-support}) and Adam-PGD leads (\cref{tab:main}).
Second, this sparse footprint interacts with heavy-tailed BEV activation skew to defeat the scalar step ($2.5\,\overline{\eps}/k$) inherited from homogeneous CP (\emph{s-50$^\dagger$}, \cref{tab:main}): with $79$--$91\%$ of post-ReLU BEV coordinates at zero background (\cref{sec:s-checks}), the tensor-wide mean $\overline{\eps}$ is vastly smaller than $\eps_i$ on active foreground cells.
On HEAL and STAMP, the $<1\%$ non-zero gradients lie exclusively on high-activation foreground anchors where $\eps_i\gg\overline{\eps}$, leaving s-50$^\dagger$ under-stepped and detections intact (\cref{fig:qual-v2x}).
While calibrating to elementwise steps ($2.5\,\eps_i/k$) saturates on-support coordinates, sign-PGD from zero leaves $>99\%$ off-support coordinates untouched.

\vspace{2pt}\noindent\textbf{Off-support neighborhood and oracle initialization.}
Masking \hp's $\delta$ to gradient support leaves $\Lrem$ unchanged yet weakens attack ($0.201\!\to\!0.326$ on HEAL, $0.704\!\to\!0.791$ on CodeFilling); off-support perturbation alone leaves AP within $0.005$ of clean (\cref{tab:s-support}).
Because adjacent anchors still predict boxes above IoU $0.5$, removing an object requires perturbing its spatial neighborhood. On HEAL/STAMP, Adam-PGD reaches lower $\Lrem$ than \hp yet higher AP (\cref{tab:s-checks}), which clamped logit loss cannot fix ($0.292$, $0.140$).
Initializing Adam-PGD from \hp's $\delta$ (\cref{tab:amort}) closes this gap on HEAL/STAMP ($0.209$, $0.066$) and matches the CodeFilling code-flip oracle ($0.688$ vs.\ $0.686$; \cref{tab:s-oracle}), confirming that search fails from zero off-support initialization.

\takeaway{Per-frame search fails because $<1\%$ gradient support leaves spatial neighborhoods unperturbed under scalar steps. Convolutional generation natively spans off-support neighborhoods, unlocking severe attack efficacy without test-time search.}

\vspace{2pt}\noindent\textbf{Per-frame generator and ablations.}
Fitting the U-Net fresh per test frame for 500 Adam steps with frame labels (\emph{Per-frame $G$}, \cref{tab:amort}) already beats Adam-PGD across sparse-support victims ($0.217$, $0.082$, $0.710$); amortization (\hp) adds a $1{,}200$--$14{,}000\times$ speedup.
Generator width, training size, or removing $\mathcal{L}_{\text{gan}}$ shifts AP by $\le0.005$ (\cref{tab:s-ablation}), while $\mathcal{L}_{\text{gan}}$ cuts patch MMD$^2$ (Squared Maximum Mean Discrepancy) to the clean manifold by $0.05$--$0.49$ (\cref{tab:s-gan-manifold}).
Universal perturbations~\cite{uap2017} and heuristic cancellations are $0.19$--$0.65$ AP weaker (\cref{tab:s-uap,tab:s-checks}).

\subsection{Deployability Analysis}
\label{sec:deploy}

\vspace{2pt}\noindent\textbf{Cross-family transfer.}
In the \emph{transfer posture} (\cref{tab:deploy}a), HEAL and STAMP generators transfer almost losslessly ($0.204$, $0.072$), and CodeFilling/UniTrans generators transfer to HEAL/STAMP ($0.109$--$0.159$).
However, no generator transfers into CodeFilling ($0.85$ vs.\ $0.876$), and transfer into UniTrans degrades ($0.564$--$0.800$).
Replaying fixed perturbations transfers poorly ($0.568$; \cref{tab:s-fixed}), whereas generators transfer reliably across continuous fusion spaces with shared gradient geometry (\cref{sec:s-transfer}).

\begin{table}[t]
\centering
\footnotesize
\setlength{\tabcolsep}{2.5pt}
\caption{\textbf{Deployability} on OPV2V-H ($\eps{=}1.0$, AP@0.5 $\downarrow$). (a) Generator transfer (last row: Adam-PGD $\delta$ replay). (b) Channel $\Delta$AP vs.\ ideal ($+$: weaker).}
\label{tab:deploy}
\begin{tabular}{@{}l|cccc@{}}
\toprule
\multicolumn{5}{@{}l}{(a) Transfer: source $\to$ victim} \\
Source & HEAL & STAMP & CodeFill. & UniTr. \\
\midrule
HEAL        & \hpcell.200 & .072 & .855 & .564 \\
STAMP       & .204 & \hpcell.063 & .847 & .656 \\
CodeFilling & .159 & .109 & \hpcell.709 & .800 \\
UniTrans    & .155 & .126 & .863 & \hpcell.128 \\
Adam $\delta$, other & .568 & -- & .860 & -- \\
\bottomrule
\end{tabular}
\\[2pt]
\begin{tabular}{@{}l|ccc@{}}
\toprule
\multicolumn{4}{@{}l}{(b) Channel: $\Delta$AP for \hp\ / sign-50} \\
Channel & HEAL & STAMP & CodeFill. \\
\midrule
1-frame delay  & +.008/+.099 & +.005/+.112 & +.004/+.030 \\
INT8           & .000/.000 & +.001/.000 & .000/.000 \\
INT4           & +.003/+.005 & $-$.008/+.002 & +.084/+.041 \\
\bottomrule
\end{tabular}
\vspace{-2mm}
\end{table}

\takeaway{Generator transfer tracks shared gradient geometry across continuous fusion spaces, but discrete codebooks structurally isolate translators from cross-family transfer.}

\vspace{2pt}\noindent\textbf{Channel delay, quantization, and unseen encoders.}
Conditioning on the ego's \emph{previous} broadcast costs \hp $\le0.008$ AP vs.\ $0.03$--$0.11$ for sign-PGD (\cref{tab:deploy}b), and lagging ego features by five frames costs $\le0.016$ AP (\cref{tab:s-history}).
INT8 quantization and mild pose noise ($0.2$\,m/$0.2^\circ$) have negligible impact, while INT4 weakens attacks on CodeFilling and GenComm without protecting HEAL or STAMP (\cref{tab:s-channel}).
On unseen HEAL collaborators (SECOND LiDAR~\cite{second18103337}, ResNet-101 camera~\cite{he2016deep}), the unchanged generator outperforms sign-5 PGD ($0.104$ vs.\ $0.239$; $0.292$ vs.\ $0.460$; \cref{tab:s-unseen}), and OPV2V-H$\to$V2X-Real GenComm transfer drops AP by $\le0.085$ (\cref{tab:s-sim2real}).
We include additional deployability result sweeps in \cref{sec:s-deploy} (\cref{tab:s-channel,tab:s-unseen,tab:s-sim2real}).

\subsection{Evaluation on V2X-Real}
\label{sec:v2xreal}

\textbf{LiDAR--LiDAR and LiDAR--camera fleets.}
On the real-world V2X-Real (\cref{tab:v2xreal}, \cref{tab:s-v2xreal,fig:qual-v2x}), simulation rankings hold: at $\eps{=}1.0$, \hp beats Adam-PGD by $0.08$--$0.09$ AP across all four HEAL/STAMP configurations at $\sim$$3{,}000\times$ lower cost, and at $\eps{=}1.5$ leads by up to $0.10$, with LiDAR--LiDAR (LL) and LiDAR--camera (LC) fleets degrading comparably.
Zeroing $\fatt$ shifts clean AP by $\le0.011$ (\cref{sec:s-v2xreal}), confirming that degradation stems from perturbation rather than missing collaborator data.

\takeaway{On real-world multi-sensor fleets, single-pass generation preserves attack efficacy at $3{,}000\times$ lower latency, proving practical vulnerability under strict 100\,ms deadlines.}

\begin{table}[t]
\centering
\footnotesize
\setlength{\tabcolsep}{2.5pt}
\caption{\textbf{V2X-Real} (vehicle AP@0.5 $\downarrow$; full sweep in \cref{tab:s-v2xreal}). LL: vehicle + roadside LiDAR; LC: LiDAR + camera. Bold: strongest shown. Seed std: \hp $\le.016$ ($\le.043$ GenComm), Adam $\le.005$ (GPU var.\ $\le.002$, \cref{sec:s-protocol}). Time/frame on Jetson AGX Thor (\hp: 2.0--2.6\,ms).}
\label{tab:v2xreal}
\begin{tabular}{@{}l|c|cc|cc@{}}
\toprule
& & \multicolumn{2}{c|}{$\eps=1.0$} & \multicolumn{2}{c}{$\eps=1.5$} \\
Victim & Clean & Adam & \hp & Adam & \hp \\
\midrule
HEAL-LL       & .744 & .372 & \hpcell\best{.290} & .185 & \hpcell\best{.149} \\
STAMP-LL      & .720 & .423 & \hpcell\best{.340} & .269 & \hpcell\best{.177} \\
CodeF.-LL     & .746 & .725 & \hpcell\best{.709} & .720 & \hpcell\best{.690} \\
GenC.-LL      & .771 & \best{.079} & \hpcell.153 & \best{.057} & \hpcell.128 \\
UniTrans-LL   & .746 & \best{.070} & \hpcell.089 & \best{.038} & \hpcell.057 \\
\midrule
HEAL-LC       & .734 & .379 & \hpcell\best{.290} & \best{.182} & \hpcell\best{.170} \\
STAMP-LC      & .728 & .430 & \hpcell\best{.340} & .312 & \hpcell\best{.210} \\
CodeF.-LC     & .720 & \best{.710} & \hpcell\best{.716} & \best{.674} & \hpcell.706 \\
GenC.-LC      & .714 & \best{.023} & \hpcell.163 & \best{.012} & \hpcell.097 \\
UniTrans-LC   & .744 & \best{.013} & \hpcell.102 & \best{.002} & \hpcell.067 \\
\midrule
s/frame & -- & 7.4--132 & \hpcell.002 & \multicolumn{2}{c}{\footnotesize (same)} \\
\bottomrule
\end{tabular}
\vspace{-2mm}
\end{table}

\vspace{2pt}\noindent\textbf{Object removal and FP suppression on V2X-Real.}
The object-level results mirror those of OPV2V-H (\cref{tab:s-object-v2x}): \hp removes $52$--$61\%$ of cleanly detected vehicles on HEAL and STAMP, $13$--$16$ percentage points more than Adam-PGD, and $85$--$87\%$ on UniTrans, yet it \emph{cuts} false positives where the per-frame baselines add them (HEAL $4.6{\to}2.0$/frame vs.\ Adam's $4.7{\to}5.3$).
Every attack erases collaborator-only objects at $80$--$100\%$, since suppressing them needs only the collaborator's own evidence; the attacks separate on objects the ego itself observes, where \hp still removes a large share (\cref{sec:s-object}), so a single poisoned peer corrupts the ego's \emph{local} perception rather than only the shared view.
We provide step sweeps and adaptive attacks in \cref{sec:s-v2xreal}, and discuss attack generality in \cref{sec:s-discussion}.

\section{Conclusion}
\label{sec:conclusion}

This paper investigates the adversarial robustness of heterogeneous cooperative perception across five translation paradigms under a matched-objective attack harness.
Diagnosing that per-frame search is bottlenecked by step-size miscalibration and a sparse ($<1\%$) gradient footprint on per-location fusion, we propose \hp: an amortized feature attack that trains a convolutional generator offline to predict spatially coherent perturbations in a single forward pass ($\sim$2\,ms on embedded hardware) without online labels, detector queries, or gradients.
Extensive experiments on OPV2V-H and V2X-Real demonstrate that \hp outperforms 500-step Adam-PGD on HEAL, STAMP, and CodeFilling at up to $14{,}000\times$ lower cost and transfers across continuous translators, latency, and sensor modalities, establishing practical vulnerabilities under strict real-time V2X deadlines.

\vspace{2pt}\noindent\textbf{Limitations and future work.}
A single forward pass does not penetrate private discrete codebooks or adapt to the online noise of stochastic diffusion decoders, where per-frame white-box search stays stronger past the real-time deadline.
Because \hp exposes a blind spot ego-only evidence cannot cover, the central next step is defenses that verify collaborator features beyond the ego's line of sight. 

\section*{LLM Usage Statement}
We used Anthropic Claude and Google Gemini to improve grammar, clarity, and concision, as well as for implementation assistance including code completion, debugging, and refactoring. No core algorithmic designs, threat models, or experimental findings were generated by LLMs. All LLM-assisted text edits and code changes were reviewed and verified by the authors for accuracy, originality, and experimental correctness.
{
    \small
    \bibliographystyle{ieeenat_fullname}
    \bibliography{references}
}

\clearpage
\def\supplincluded{}
\newcommand{\mref}[1]{\ref{#1}}
\ifx\supplincluded\undefined
  \documentclass[10pt,twocolumn,letterpaper]{article}
  \usepackage[review]{cvpr}
  
  \usepackage{xr-hyper}
  \externaldocument[M-]{main}
  \definecolor{cvprblue}{rgb}{0.21,0.49,0.74}
  \usepackage[pagebackref,breaklinks,colorlinks,allcolors=cvprblue]{hyperref}
  \usepackage[capitalize]{cleveref}
  \crefname{section}{Sec.}{Secs.}
  \crefname{table}{Tab.}{Tabs.}
  \def\paperID{*****}
  \def\confName{CVPR}
  \def\confYear{2027}
  \title{From Per-Frame Search to Practical Attacks: Investigating Adversarial Robustness on Multi-Agent Heterogeneous Cooperative Perception}
  \author{Anonymous CVPR submission}
  \newcommand{\mref}[1]{\ref{M-#1}}
  \begin{document}
\fi

\setcounter{section}{0}
\renewcommand{\thesection}{S\arabic{section}}
\renewcommand{\theHsection}{S\arabic{section}}
\setcounter{table}{0}
\renewcommand{\thetable}{S\arabic{table}}
\renewcommand{\theHtable}{S\arabic{table}}
\setcounter{figure}{0}
\renewcommand{\thefigure}{S\arabic{figure}}
\renewcommand{\theHfigure}{S\arabic{figure}}
\maketitlesupplementary

\renewcommand{\topfraction}{0.92}
\renewcommand{\bottomfraction}{0.85}
\renewcommand{\textfraction}{0.08}
\renewcommand{\floatpagefraction}{0.75}
\renewcommand{\dbltopfraction}{0.92}
\renewcommand{\dblfloatpagefraction}{0.75}
\setcounter{topnumber}{3}
\setcounter{bottomnumber}{2}
\setcounter{totalnumber}{3}
\setcounter{dbltopnumber}{3}
\makeatletter
\setlength{\@fptop}{0pt}
\setlength{\@fpsep}{12pt}
\setlength{\@fpbot}{0pt plus 1fil}
\makeatother

This supplement provides the evaluation protocol (\cref{sec:s-protocol}), \hp implementation details (\cref{sec:s-impl}), and complete result tables supporting the main paper: OPV2V-H (\cref{sec:s-opv2v}), per-frame baselines and gradient footprint (\cref{sec:s-checks}), ablations (\cref{sec:s-ablation}), object-level metrics (\cref{sec:s-object}), transfer and controls (\cref{sec:s-transfer}), V2X-Real (\cref{sec:s-v2xreal}), deployability (\cref{sec:s-deploy}), discussion and published defenses (\cref{sec:s-discussion}), and qualitative figure provenance (\cref{sec:s-qual}).
Tables, sections, and figures in the main paper are referenced by their main-paper numbers (\eg, main Tab.~\mref{tab:main}).
Unless stated otherwise, values are AP@0.5 after attack (lower indicates a stronger attack), $\pm$ denotes the sample standard deviation over three seeds, and bold / underline denote the strongest / weakest attack (excluding Clean and Floor).

\section{Evaluation Protocol}
\label{sec:s-protocol}

\textbf{Protocol and frame sets.}
A versioned protocol file fixes, per victim, the model directory and pinned checkpoint, evaluation range, frame sets, attacker selection rule, perturbation budgets, and generator checkpoints (one per victim, budget, and seed).
Every result file records the protocol hash, frame-set hash, number of frames expected and processed, and a completeness flag; the table generator rejects incomplete runs and mismatched hashes.
No number in the paper was transcribed manually from logs.
Frame sets are all test frames with at least one collaborator.

\vspace{2pt}\noindent\textbf{Perturbation budget.}
Every attack, including the random floor and \hp, uses the elementwise bound of main Eq.~(\mref{eq:budget}), computed from the ego's feature in that frame.
For GenComm, the bound is computed on the two-channel diffusion message.

\vspace{2pt}\noindent\textbf{Per-frame baselines.}
All per-frame attacks minimize the removal loss on the victim's real detection heads using the frame's ground-truth labels, under the victim's true translation and fusion forward path.
sign-PGD-$k$ takes $k$ steps of elementwise size $2.5\,\eps_i/k$ from zero and returns the final iterate.
Adam-PGD uses moments $(0.9, 0.999)$ without bias correction and the same elementwise step size across 5 restarts of 100 steps; the first restart initializes at zero and the remaining four draw uniformly from the ball, retaining the restart with the lowest final loss.
EOT-8 averages gradients (and restart-selection losses) over 8 stochastic forward passes.
BPDA replaces CodeFilling's hard code assignment in the backward pass with a softmax over codebook distances at temperature $\tau$.
Because few-step PGD with step size $2.5\,\eps_i/k$ can overshoot at $k{=}2$ ($1.25\,\eps_i$), sign-2 is not monotone between sign-1 and sign-5; we include it in the complete tables and use sign-1 and sign-5 as the matched-compute baselines.

\vspace{2pt}\noindent\textbf{Scalar step-size baseline.}
Prior homogeneous CP attacks~\cite{tu2021advcomm,datafab2024,sombra2025} use a single scalar step size $\alpha = 2.5\,\mathrm{mean}(\eps_i)/k$ across the feature tensor.
Under our elementwise relative budget (main Eq.~\mref{eq:budget}), $79$--$91\%$ of post-ReLU BEV coordinates are zero background (\cref{sec:s-checks}), dragging $\mathrm{mean}(\eps_i)$ far below the local bound $\eps_i$ on active foreground object pillars.
When downstream spatial networks spread gradients across the BEV grid ($76.5\%$ support on homogeneous V2VAM and $100\%$ on UniTrans; \cref{tab:s-support}), scalar-step sign-PGD-50 still degrades detection substantially (dropping V2VAM from $0.957$ to $0.630$ and UniTrans from $0.882$ to $0.299$ at $\eps{=}1.0$; \cref{tab:s-full}).
On sparse-support heterogeneous victims (HEAL and STAMP, where gradients are non-zero on only $0.66$--$0.73\%$ of coordinates), however, those non-zero gradients lie exclusively on high-activation foreground anchors where $\eps_i \gg \mathrm{mean}(\eps_i)$.
Consequently, scalar-step sign-PGD-50 updates $<1\%$ of coordinates by at most $2.5\,\mathrm{mean}(\eps_i) \ll \eps_i$ and leaves HEAL at $0.822$ and STAMP at $0.727$ ($\eps{=}1.0$)---weaker than the uniform random-$\delta$ floor ($0.744$ and $0.716$)---illustrating how a standard homogeneous CP attack configuration makes heterogeneous CP appear artificially robust.
Upgrading to the elementwise step $2.5\,\eps_i/k$ drops HEAL to $0.373$, STAMP to $0.217$, and V2VAM to $0.081$; we report ``sign-PGD-50 scalar'' ($^\dagger$ in main Tab.~\mref{tab:main}) in \cref{tab:s-full} and use the calibrated elementwise step everywhere else.

\vspace{2pt}\noindent\textbf{Attacker selection.}
On OPV2V-H, the attacker is the first collaborator in dataset order; on V2X-Real, it is the first collaborator whose encoder differs from the ego's.
The same collaborator is attacked across all methods on any given frame.
Frames without a collaborator are not attacked and are excluded from the evaluation sets.

\vspace{2pt}\noindent\textbf{Latency and offline training cost.}
Crafting times (main Tab.~\mref{tab:main}, main Fig.~\mref{fig:cost}, and \cref{tab:s-latency}) are median CUDA-synchronized milliseconds to produce $\delta$ for a single frame on an NVIDIA Jetson AGX Thor (an embedded automotive GPU), measured in a single process using cached features with the first two frames dropped as warm-up.
On this platform, \hp takes $2.0$\,ms per frame across all five heterogeneous victims ($18$\,ms on V2VAM), whereas 1-step sign-PGD requires $6.1$--$81.3$\,ms ($121$\,ms on V2VAM) and 500-step Adam-PGD requires $2.3$ to $28.6$\,s per frame ($41.4$\,s on V2VAM; \cref{tab:s-latency}).
Training one generator ($1{,}000$ updates) takes $6$--$16$ GPU-minutes (e.g., STAMP 7, CodeFilling 12, HEAL 15, GenComm 15).
Relative to Adam-PGD's on-device per-frame latency, this one-time offline cost is amortized after just $70$--$390$ attacked frames ($7$--$39$\,s of driving at 10\,Hz), whereas per-frame attacks incur their full optimization latency and require ground-truth labels on every frame.

\begin{table}[tb]
\centering\small
\setlength{\tabcolsep}{3.5pt}
\caption{\textbf{Per-victim crafting latency on NVIDIA Jetson AGX Thor} (median ms / frame on OPV2V-H, $\eps{=}1.0$; main Tab.~\mref{tab:main} and main Fig.~\mref{fig:cost}). Bold / underline: fastest / slowest attack per victim. Fus.: the attack-independent fusion$+$detection-head forward that runs every frame downstream of the perturbation (same device, cached features); a frame's attacked cost is encoder $+$ craft $+$ Fus.}
\label{tab:s-latency}
\resizebox{\linewidth}{!}{%
\begin{tabular}{@{}l|ccccc|c|c@{}}
\toprule
Victim & s-1 & s-5 & s-50 & s-400 & Adam & \hp & Fus. \\
\midrule
HEAL        & 6.1 & 24.2 & 224 & 1{,}776 & \underline{2{,}304} & \hpcell\best{2.0} & 1.7 \\
STAMP       & 7.4 & 28.7 & 269 & 2{,}125 & \underline{2{,}759} & \hpcell\best{2.0} & 1.7 \\
CodeFilling & 19.6 & 72.5 & 667 & 5{,}294 & \underline{6{,}732} & \hpcell\best{2.0} & 9.2 \\
GenComm     & 41.2 & 140.1 & 1{,}311 & 10{,}277 & \underline{13{,}023} & \hpcell\best{2.0} & 15.7 \\
UniTrans    & 81.3 & 309.6 & 2{,}895 & 22{,}946 & \underline{28{,}624} & \hpcell\best{2.0} & 41.9 \\
\midrule
V2VAM (homo.) & 121.0 & -- & 4{,}000 & -- & \underline{41{,}400} & \hpcell\best{18.0} & -- \\
\bottomrule
\end{tabular}}
\end{table}

\vspace{2pt}\noindent\textbf{Crafting is not the whole frame.}
The times above are crafting only: a crafted perturbation must still traverse fusion and the detection heads before the frame produces output, and this downstream pass (Fus.\ column, attack-independent) costs $1.7$--$41.9$\,ms on Thor, growing with translator complexity ($41.9$\,ms for UniTrans's transformer fusion).
An attacked frame therefore costs encoder $+$ craft $+$ fusion/heads.
UniTrans s-1, whose $81.3$\,ms craft alone sits under the $100$\,ms budget, yields a $123$\,ms frame once its $41.9$\,ms fusion pass is added, exceeding the deadline \emph{before} the ego and collaborator encoders are counted; because crafting cost is step-count- rather than budget-bound, the same $123$\,ms holds at $\eps{=}1.5$, where s-1 is stronger, so it too is deadline-infeasible.
\hp's $2$\,ms craft instead leaves the budget to perception.

\vspace{2pt}\noindent\textbf{Reproducibility.}
LiDAR inference with sparse convolutions is not bitwise deterministic on GPUs: rerunning a clean evaluation changes AP@0.5 by at most $0.0005$ on V2X-Real and $0.0001$ on OPV2V-H.
We report all AP metrics to three decimal places.

\section{\hp Implementation}
\label{sec:s-impl}

\textbf{Generator architecture.}
$G_\theta$ is a three-scale U-Net.
The context $c\in\mathbb{R}^{3C\times H\times W}$ passes through three Conv--InstanceNorm--ReLU blocks ($3\times3$; $3C\!\to\!64$ at stride 1, $64\!\to\!128$ and $128\!\to\!256$ at stride 2), three residual blocks at 256 channels, and a decoder with two $4\times4$ transposed convolutions and skip connections, followed by a $3\times3$ output convolution to $C$ channels and a $\tanh$ activation.
Decoder activations are bilinearly resized to match skip tensors before concatenation, allowing a single architecture to accept arbitrary BEV grid dimensions.
The channel dimension is $C{=}128$ for HEAL, STAMP, CodeFilling, and UniTrans on OPV2V-H ($C{=}256$ on V2X-Real), and $C{=}2$ for GenComm's diffusion message.
The generator has $5.07$\,M parameters at $C{=}128$ and $4.78$\,M at $C{=}2$.

\vspace{2pt}\noindent\textbf{Discriminator architecture.}
We use a four-layer PatchGAN discriminator ($C\!\to\!64\!\to\!128\!\to\!256\!\to\!1$), with stride 2 and LeakyReLU(0.2) on the first three layers and InstanceNorm on layers 2--3 ($0.79$\,M parameters at $C{=}128$).
The discriminator is used only during offline training and discarded at deployment.

\vspace{2pt}\noindent\textbf{Training objective.}
The training loss is $\lambda_{\text{rem}}\Lrem + \lambda_{\text{gan}}\mathcal{L}_{\text{gan}} + \lambda_{\text{hinge}}\mathcal{L}_{\text{hinge}}$ with weights $(\lambda_{\text{rem}},\lambda_{\text{gan}},\lambda_{\text{hinge}}) = (10, 0.1, 1)$ fixed across all victims.
Here $\mathcal{L}_{\text{gan}} = \tfrac12\mathbb{E}[(D(\fatt+\delta)-1)^2]$ is trained against the least-squares discriminator loss $\tfrac12\mathbb{E}[(D(\fatt)-1)^2]+\tfrac12\mathbb{E}[D(\fatt+\delta)^2]$ with one $D$ update per $G$ update, and $\mathcal{L}_{\text{hinge}} = \mathrm{ReLU}(\|\delta\|_2 - \|\eps(x)\|_2)$ is an auxiliary norm guard that evaluates to zero by construction under main Eq.~(\mref{eq:delta}).

\vspace{2pt}\noindent\textbf{Optimization recipe.}
We train $G_\theta$ using Adam ($\beta=(0.5,0.999)$, learning rate $2\times10^{-4}$, no weight decay or schedule) for $1{,}000$ single-frame updates drawn from one shuffled pass over the training split (skipping frames without a collaborator; the data-size ablation restricts this pool to 100 or 300 frames), with the surrogate victim stack frozen in evaluation mode.
All generators are trained strictly on the training split of each dataset (never on test frames), using three seeds (0, 1, 2) per victim and budget.
One shared recipe is used for every victim; a GenComm-tuned weighting ($\lambda_{\text{gan}}{=}1$, $\lambda_{\text{hinge}}{=}10$) reaches $0.613$ at $\eps{=}1.5$ (seed 0) vs.\ $0.673$ for the shared recipe on the same seed (three-seed mean $0.655\pm0.031$). Because GenComm's stochastic diffusion decoder induces higher training variance across seeds ($\pm0.031$--$0.049$), differences among \hp recipe variants on GenComm are not statistically significant and do not alter any comparative conclusion against per-frame baselines.

\vspace{2pt}\noindent\textbf{Multi-sample diffusion training.}
For GenComm, we also trained generators that average the training loss over 4 diffusion-noise draws per update (``\hp-EOT4'' in \cref{tab:s-full,tab:s-adaptive}); this variant yields comparable performance within seed variance.

\section{Complete OPV2V-H Results}
\label{sec:s-opv2v}

\begin{table*}[tp]
\centering\small
\setlength{\tabcolsep}{4pt}
\caption{\textbf{OPV2V-H results with matched harness} (AP@0.5 $\downarrow$) under different perturbation budgets. sign-PGD-50 scalar ($^\dagger$ in main Tab.~\mref{tab:main}) is the legacy scalar-step sign-PGD (\cref{sec:s-protocol}). \hp-EOT4 averages four diffusion draws and applies only to GenComm, the one stochastic victim (--: not applicable). Bold / underline: strongest / weakest attack (excluding Floor).}
\label{tab:s-full}
\begin{tabular}{@{}l|l|cc|ccccc|cc@{}}
\toprule
$\eps$ & Victim & Clean & Floor & s-1 & s-2 & s-5 & s-50 & sign-PGD-50 scalar & \hp & \hp-EOT4 \\
\midrule
\multirow{6}{*}{1.0}
& HEAL        & .866 & .744 & .374 & .604 & .381 & .373 & \underline{.822} & \hpcell\best{.200}\pmstd{.000} & -- \\
& STAMP       & .899 & .716 & .255 & .450 & .224 & .217 & \underline{.727} & \hpcell\best{.063}\pmstd{.007} & -- \\
& CodeFilling & .876 & .873 & \underline{.863} & .849 & .825 & .803 & .848 & \hpcell\best{.709}\pmstd{.002} & -- \\
& GenComm     & .880 & .877 & \underline{.863} & .856 & .813 & \best{.630} & .785 & \hpcell.768\pmstd{.049} & \hpcell.757\pmstd{.067} \\
& UniTrans    & .882 & .866 & .263 & \underline{.468} & .171 & .288 & .299 & \hpcell\best{.128}\pmstd{.024} & -- \\
& V2VAM       & .957 & .946 & .417 & .148 & .060 & .081 & \underline{.630} & \hpcell\best{.043}\pmstd{.002} & -- \\
\midrule
\multirow{6}{*}{1.5}
& HEAL        & .866 & .400 & .279 & .602 & .306 & .293 & \underline{.754} & \hpcell\best{.130}\pmstd{.000} & -- \\
& STAMP       & .899 & .342 & .189 & .401 & .176 & .158 & \underline{.537} & \hpcell\best{.032}\pmstd{.001} & -- \\
& CodeFilling & .876 & .872 & \underline{.857} & .840 & .816 & .786 & .823 & \hpcell\best{.646}\pmstd{.002} & -- \\
& GenComm     & .880 & .874 & \underline{.858} & .846 & .747 & \best{.395} & .619 & \hpcell.655\pmstd{.031} & \hpcell.648\pmstd{.011} \\
& UniTrans    & .882 & .757 & \best{.038} & .274 & .042 & .201 & \underline{.289} & \hpcell.078\pmstd{.013} & -- \\
& V2VAM       & .957 & .929 & .208 & .046 & .018 & .040 & \underline{.447} & \hpcell\best{.017}\pmstd{.001} & -- \\
\bottomrule
\end{tabular}
\end{table*}

\begin{table*}[tp]
\centering\small
\setlength{\tabcolsep}{3.6pt}
\caption{\textbf{Adaptive attacks on CodeFilling and GenComm on OPV2V-H} (AP@0.5 $\downarrow$). EOT-8 and BPDA target the stochastic (GenComm) and non-differentiable (CodeFilling) translations; BPDA applies only to CodeFilling's code assignment, and \hp-EOT4 applies only to GenComm (--: not applicable). Bold / underline: strongest / weakest attack (excluding Floor).}
\label{tab:s-adaptive}
\resizebox{\linewidth}{!}{%
\begin{tabular}{@{}l|l|cc|ccccccc|ccc|cc@{}}
\toprule
$\eps$ & Victim & Clean & Floor & s-1 & s-2 & s-5 & s-50 & s-400 & Adam & EOT-8 & \multicolumn{3}{c|}{BPDA $\tau{=}$0.1 / 0.5 / 1} & \hp & \hp-EOT4 \\
\midrule
\multirow{2}{*}{1.0}
& CodeFilling & .876 & .873 & \underline{.863} & .848 & .824 & .803 & .801 & .800 & .821 & .859 & .840 & .832 & \hpcell\best{.709}\pmstd{.002} & -- \\
& GenComm     & .880 & .877 & \underline{.863} & .856 & .813 & .630 & .595 & \best{.369} & .382 & -- & -- & -- & \hpcell.768\pmstd{.049} & \hpcell.756\pmstd{.067} \\
\midrule
\multirow{2}{*}{1.5}
& CodeFilling & .876 & .872 & \underline{.857} & .839 & .816 & .786 & .783 & .784 & .813 & \underline{.857} & .836 & .824 & \hpcell\best{.646}\pmstd{.002} & -- \\
& GenComm     & .880 & .874 & \underline{.858} & .846 & .746 & .395 & .301 & .076 & \best{.072} & -- & -- & -- & \hpcell.655\pmstd{.031} & \hpcell.647\pmstd{.012} \\
\bottomrule
\end{tabular}}
\end{table*}

\begin{table}[tb]
\centering\small
\setlength{\tabcolsep}{4pt}
\caption{\textbf{Seed variance on OPV2V-H}. Adam-PGD: three initialization seeds; \hp: three training seeds.}
\label{tab:s-var}
\begin{tabular}{@{}l|cc|cc@{}}
\toprule
& \multicolumn{2}{c|}{$\eps=1.0$} & \multicolumn{2}{c}{$\eps=1.5$} \\
Victim & Adam & \hp & Adam & \hp \\
\midrule
HEAL        & \underline{.271}\pmstd{.002} & \hpcell\best{.201}\pmstd{.000} & \best{.088}\pmstd{.002} & \hpcell\underline{.130}\pmstd{.001} \\
STAMP       & \underline{.133}\pmstd{.005} & \hpcell\best{.063}\pmstd{.008} & \best{.022}\pmstd{.001} & \hpcell\underline{.033}\pmstd{.002} \\
CodeFilling & \underline{.801}\pmstd{.002} & \hpcell\best{.706}\pmstd{.002} & \underline{.785}\pmstd{.002} & \hpcell\best{.644}\pmstd{.001} \\
GenComm     & \best{.369}\pmstd{.003} & \hpcell\underline{.776}\pmstd{.047} & \best{.078}\pmstd{.003} & \hpcell\underline{.657}\pmstd{.030} \\
UniTrans    & \best{.064}\pmstd{.002} & \hpcell\underline{.131}\pmstd{.024} & \best{.016}\pmstd{.002} & \hpcell\underline{.081}\pmstd{.012} \\
V2VAM       & \best{.027}\pmstd{.004} & \hpcell\underline{.042}\pmstd{.002} & \best{.011}\pmstd{.002} & \hpcell\underline{.018}\pmstd{.000} \\
\bottomrule
\end{tabular}
\end{table}

\begin{table}[tb]
\centering\small
\setlength{\tabcolsep}{3.5pt}
\caption{\textbf{AP@0.7} on OPV2V-H, the stricter threshold of the HEAL/STAMP evaluations; every ranking of main Tab.~\mref{tab:main} at AP@0.5 is unchanged.}
\label{tab:s-ap07}
\setlength{\tabcolsep}{1.5pt}
\begin{tabular}{@{}l|c|ccc|cc@{}}
\toprule
& & \multicolumn{3}{c|}{$\eps=1.0$} & \multicolumn{2}{c}{$\eps=1.5$} \\
Victim & Clean & s-50 & Adam & \hp & Adam & \hp \\
\midrule
HEAL        & .685 & \underline{.240} & .185 & \hpcell\best{.148}\pmstd{.000} & \best{.059} & \hpcell\underline{.102}\pmstd{.000} \\
STAMP       & .707 & \underline{.122} & .072 & \hpcell\best{.042}\pmstd{.005} & \best{.010} & \hpcell\underline{.020}\pmstd{.001} \\
CodeFilling & .703 & \underline{.622} & \underline{.622} & \hpcell\best{.590}\pmstd{.001} & \underline{.604} & \hpcell\best{.538}\pmstd{.001} \\
GenComm     & .682 & .454 & \best{.237} & \hpcell\underline{.577}\pmstd{.039} & \best{.035} & \hpcell\underline{.479}\pmstd{.029} \\
UniTrans    & .710 & \underline{.171} & \best{.047} & \hpcell.106\pmstd{.020} & \best{.011} & \hpcell\underline{.059}\pmstd{.008} \\
V2VAM       & .921 & .010 & \best{.004} & \hpcell\underline{.035}\pmstd{.001} & \best{.001} & \hpcell\underline{.013}\pmstd{.000} \\
\bottomrule
\end{tabular}
\end{table}

\textbf{Seed variance and AP@0.7} (\cref{tab:s-var,tab:s-ap07}).
Every comparison in main Tab.~\mref{tab:main} is resolved well beyond seed noise: across all six victims (including UniTrans, where \hp std is $0.024$ at $\eps{=}1.0$ and $0.012$ at $\eps{=}1.5$, vs.\ $\le0.002$ for Adam-PGD), the smallest gap between Adam-PGD and \hp (V2VAM at $\eps{=}1.5$, $0.007$) is at least $3.5\times$ the larger of the two standard deviations.
At the stricter AP@0.7 threshold (\cref{tab:s-ap07}), every pairwise ranking from AP@0.5 holds unchanged across all five heterogeneous victims and V2VAM.

\vspace{2pt}\noindent\textbf{Adaptive white-box attacks (EOT-8 and BPDA)} (\cref{tab:s-adaptive}).
To test whether stochastic or discrete translators mask gradients, we evaluate EOT-8~\cite{athalye2018eot} (averaging gradients over 8 stochastic passes on CodeFilling Gumbel sampling and GenComm diffusion noise) and BPDA~\cite{athalye2018obfuscated} (replacing CodeFilling's hard nearest-neighbor lookup with a softmax backward pass at $\tau\in\{0.1,0.5,1\}$).
On CodeFilling, neither EOT-8 ($0.821$ at $\eps{=}1.0$, $0.813$ at $\eps{=}1.5$) nor BPDA across temperatures ($0.832$--$0.859$ at $\eps{=}1.0$, $0.824$--$0.857$ at $\eps{=}1.5$) improves over straight-through Adam-PGD ($0.800$, $0.784$), because CodeFilling's bottleneck is the sparse ($0.71\%$) spatial gradient footprint rather than the straight-through surrogate.
On GenComm, EOT-8 closely tracks Adam-PGD ($0.382$ vs.\ $0.369$ at $\eps{=}1.0$; $0.072$ vs.\ $0.076$ at $\eps{=}1.5$), while averaging 4 diffusion draws during \hp training (\hp-EOT4: $0.756$, $0.647$) remains within seed variance of single-sample \hp ($0.768$, $0.655$).

\vspace{2pt}\noindent\textbf{Larger budgets and codeword flips on CodeFilling} (\cref{tab:s-flip}).
Raising Adam-PGD's budget to $\eps\in\{2,3,5\}$ (clean $0.874$) leaves CodeFilling at $0.814$, $0.816$, and $0.812$ (whereas the same sweep collapses GenComm to $0.047$, $0.025$, and $0.018$), because extra budget is allocated to coordinates where the per-frame gradient is identically zero (\cref{sec:s-checks}).
Quantizing the attacker's feature with and without each attack's $\delta$ (\cref{tab:s-flip}) shows why: uniform random noise and Adam-PGD fill the ball to the same extent and flip nearly identical shares of codewords across victims and budgets ($14.6\%\!\to\!22.7\%$ for random on OPV2V-H, with Adam-PGD within $1.0$ percentage point).
Because Adam-PGD updates only the $<1\%$ of coordinates with a non-zero gradient and leaves the rest at its uniform random initialization, its codeword flips are predominantly random, and its advantage over random noise lies in \emph{which} codewords it flips ($0.800$ vs.\ $0.873$ at $\eps{=}1.0$) rather than how many.
By contrast, \hp flips four to five times as many codewords on OPV2V-H ($72.6\%$ and $89.0\%$), and quantization amplifies rather than attenuates its perturbation.
On V2X-Real, \hp's flip rate remains near $30\%$ at both budgets; on LC, where silencing the attacker leaves ego AP unchanged (main Sec.~\mref{sec:v2xreal}), random noise and Adam-PGD lower AP slightly at $\eps{=}1.5$ ($0.680$ and $0.674$) with one-third of the flips by corrupting fused features rather than suppressing collaborator evidence.

\begin{table*}[tp]
\centering\small
\setlength{\tabcolsep}{4pt}
\caption{\textbf{What each attack does to CodeFilling's codebook} (\hp over three seeds). Flip: share of the attacker's code indices that change. Fill: mean $|\delta_i|/\eps_i$. Zero: share of coordinates with $\delta_i{=}0$. Pass: $\lVert Q(f{+}\delta){-}Q(f)\rVert/\lVert\delta\rVert$. AP@0.5 ($\downarrow$) from the main runs (bold / underline: strongest / weakest attack excluding random).}
\label{tab:s-flip}
\begin{tabular}{@{}ll|ccccc|ccccc@{}}
\toprule
&& \multicolumn{5}{c|}{$\eps=1.0$} & \multicolumn{5}{c}{$\eps=1.5$} \\
& Attack & Flip & Fill & Zero & Pass & AP & Flip & Fill & Zero & Pass & AP \\
\midrule
\multirow{4}{*}{OPV2V-H} & random & 14.6\% & .50 & .00 & 0.60 & .873 & 22.7\% & .50 & .00 & 0.57 & .872 \\
 & sign-50 & 0.5\% & .01 & .99 & 2.43 & \underline{.803} & 0.6\% & .01 & .99 & 1.85 & \underline{.786} \\
 & Adam & 14.0\% & .47 & .07 & 0.73 & .800 & 21.7\% & .47 & .06 & 0.65 & \underline{.784} \\
 & \hp & 72.6\% & .89 & .00 & 1.56 & \hpcell\best{.709} & 89.0\% & .88 & .00 & 1.27 & \hpcell\best{.646} \\
\midrule
\multirow{4}{*}{V2X-R LL} & random & 7.5\% & .50 & .00 & 0.59 & .744 & 12.0\% & .50 & .00 & 0.61 & .739 \\
 & sign-50 & 0.7\% & .01 & .98 & 1.84 & .723 & 0.7\% & .01 & .98 & 1.26 & .719 \\
 & Adam & 7.4\% & .48 & .05 & 0.68 & \underline{.725} & 11.5\% & .47 & .06 & 0.65 & \underline{.720} \\
 & \hp & 29.9\% & .67 & .00 & 1.78 & \hpcell\best{.709} & 32.4\% & .65 & .00 & 1.30 & \hpcell\best{.690} \\
\midrule
\multirow{4}{*}{V2X-R LC} & random & 5.6\% & .50 & .00 & 0.71 & .712 & 10.7\% & .50 & .00 & 0.70 & .680 \\
 & sign-50 & 0.6\% & .02 & .98 & 1.70 & \underline{.718} & 0.7\% & .02 & .98 & 1.26 & \underline{.715} \\
 & Adam & 5.9\% & .49 & .02 & 0.77 & \best{.710} & 10.8\% & .49 & .03 & 0.72 & \best{.674} \\
 & \hp & 30.0\% & .70 & .00 & 1.33 & \hpcell.716 & 30.3\% & .65 & .00 & 1.01 & \hpcell.706 \\
\bottomrule
\end{tabular}
\end{table*}

\begin{table}[tb]
\centering\small
\caption{\textbf{CodeFilling code-channel ceiling} (OPV2V-H, AP@0.5\,$\downarrow$; \hp over three seeds). The code-flip oracle may transmit \emph{any} codebook indices within a Hamming budget (a fraction of the attacker's active cells), upper-bounding the damage reachable through the discrete channel; it is a calibration measurement, not an attack within our threat model. Soft-anneal: the \hp generator trained against a temperature-annealed soft quantizer and evaluated against the real hard codebook.}
\label{tab:s-oracle}
\begin{tabular}{@{}l|c@{}}
\toprule
 & AP@0.5 \\
\midrule
Clean & .876 \\
\hp (full recipe) & \hpcell.707\pmstd{.001} \\
\hp (soft-anneal codebook) & \hpcell.706\pmstd{.002} \\
\midrule
Code oracle, $10\%$ budget & \underline{.748} \\
Code oracle, $25\%$ budget & .711 \\
Code oracle, $50\%$ budget & .696 \\
Code oracle, unconstrained & \best{.686}\pmstd{.001} \\
\bottomrule
\end{tabular}
\end{table}

\vspace{2pt}\noindent\textbf{Code-channel ceiling and soft-anneal training} (\cref{tab:s-oracle}).
Whereas \cref{tab:s-flip} counts how many codewords \hp flips, \cref{tab:s-oracle} measures how much damage CodeFilling's discrete channel can carry at all using a code-flip oracle that transmits \emph{any} codebook indices within a Hamming budget (a fraction of the attacker's active cells).
This oracle ignores the $\eps$-ball and continuous-feature interface to upper-bound the AP drop reachable through the discrete channel.
Even unconstrained, the oracle reaches only $0.686$, just $0.021$ below \hp's $0.707$: \hp already inflicts $\sim$$89\%$ of the damage any code-level attacker could ($0.169$ of the $0.190$ AP drop from clean $0.876$), and freely flipping a quarter of the attacker's codes ($0.711$) merely matches \hp's $\eps$-bounded single pass.
The ceiling itself stays far above HEAL and STAMP ($0.201$, $0.062$) under identical ego fusion and detection heads, because the discrete channel lets the attacker at most suppress its own contribution: the codebook bounds \emph{every} attacker, not only gradient-based ones (main Sec.~\mref{sec:why}).
Consistent with an information limit rather than an optimization failure, training the \hp generator against a temperature-annealed soft quantizer---a dense surrogate gradient annealed to the hard quantizer and evaluated on the real hard codebook---leaves AP unchanged within seed variance ($0.706\pm0.002$ vs.\ $0.707\pm0.001$).

\begin{figure*}[tp]
\centering
\includegraphics[width=0.92\textwidth]{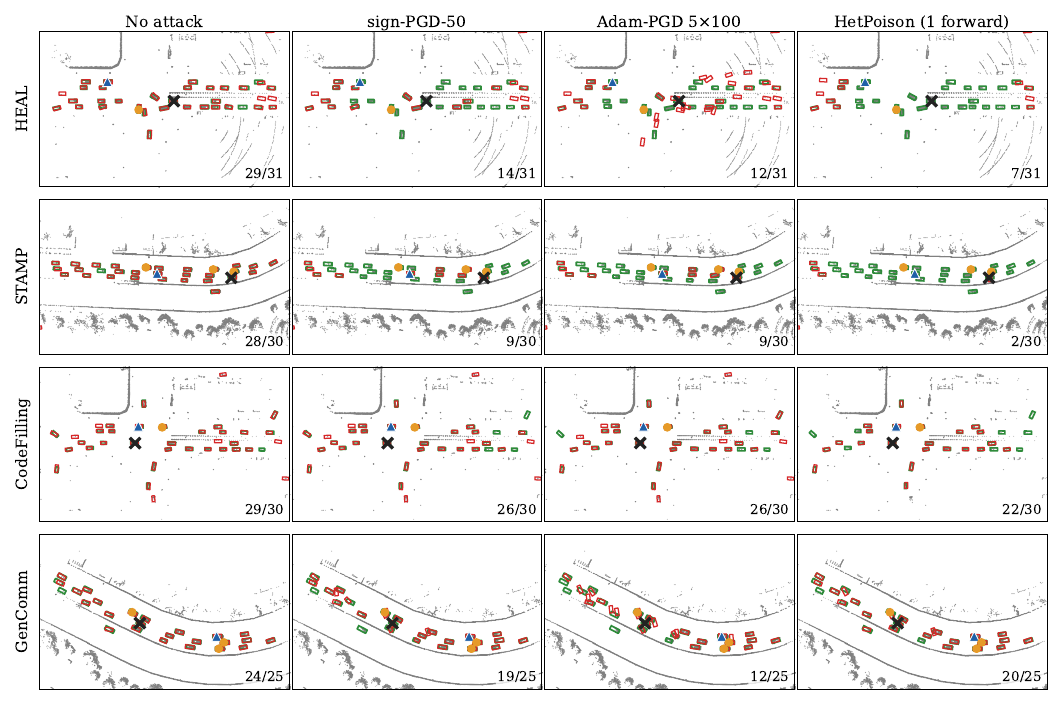}
\vspace{-3mm}
\caption{\textbf{Qualitative visualization on OPV2V-H} ($\eps{=}1.0$; seed 0 for \hp). Green: ground truth; red: fused detections; {\color[RGB]{31,95,168}$\blacktriangle$} ego; $\times$ attacker; {\color[RGB]{230,154,40}$\bullet$} benign collaborator; gray: fused LiDAR points. Numbers: objects detected (IoU $\ge 0.5$) / ground truth. Frames are selected automatically as the test frame with per-attack recalls closest to test-set means ($\ge24$ objects, clean recall $\ge0.85$; \cref{sec:s-qual}).}
\label{fig:qual}
\vspace{-2mm}
\end{figure*}

\vspace{2pt}\noindent\textbf{Qualitative visualization on OPV2V-H} (\cref{fig:qual}).
\Cref{fig:qual} visualizes fused BEV detections across HEAL, STAMP, CodeFilling, and GenComm on OPV2V-H ($\eps{=}1.0$).
On HEAL and STAMP, \hp erases both collaborator-observed and ego-visible vehicles in a single forward pass while avoiding the false-positive clutter produced by 500-step Adam-PGD; on CodeFilling, \hp removes distant objects where per-frame sign-PGD-50 and Adam-PGD leave detections intact; and on GenComm, 500-step Adam-PGD exploits dense diffusion gradients to remove more objects than single-pass \hp.

\vspace{2pt}\noindent\textbf{Train--test split isolation.}
All generators reported in this paper are trained exclusively on the training split.
Compared against an earlier prototype trained on test-split frames, training strictly on the training split changes AP on HEAL, STAMP, and CodeFilling by at most $0.009$, confirming that \hp does not rely on test-distribution exposure.

\section{Per-Frame Baselines and Gradient Footprint}
\label{sec:s-checks}

\begin{table*}[tp]
\centering\small
\setlength{\tabcolsep}{4.5pt}
\caption{\textbf{Stronger per-frame baselines and simple heuristics} (OPV2V-H, $\eps{=}1.0$, \hp seed 0). Adam-PGD variants use 5 restarts $\times$ 100 steps. Logit: removal loss on the positive anchors' logits clamped at $-5$. From \hp: first restart at \hp's $\delta$. $-\eps\,\mathrm{sign}$, cancel: $\delta{=}{-}\eps\odot\mathrm{sign}(\fatt)$ and $\delta{=}{-}\mathrm{clip}(\fatt,\pm\eps)$. $\lVert\delta\rVert$: relative to $\lVert\fatt\rVert$. Bold / underline: strongest / weakest attack ($^\dagger$oracle start, excluded from bolding). --: undefined for the clean row ($\delta{=}0$).}
\label{tab:s-checks}
\begin{tabular}{@{}l|ccc|ccc|ccc@{}}
\toprule
& \multicolumn{3}{c|}{HEAL} & \multicolumn{3}{c|}{STAMP} & \multicolumn{3}{c}{CodeFilling} \\
Attack & AP & $\Lrem$ & $\lVert\delta\rVert$ & AP & $\Lrem$ & $\lVert\delta\rVert$ & AP & $\Lrem$ & $\lVert\delta\rVert$ \\
\midrule
Clean               & .866 & .616 & -- & .899 & .572 & -- & .876 & .629 & -- \\
Adam-PGD            & .275 & .132 & .27 & .127 & .047 & .45 & .799 & .497 & .95 \\
\quad logit loss    & .292 & .132 & .21 & .140 & .047 & .29 & .796 & .493 & .94 \\
\quad from \hp's $\delta^\dagger$ & .209 & .132 & .53 & .066 & .046 & .98 & .688 & .467 & 1.50 \\
$-\eps\,\mathrm{sign}(\fatt)$ & \underline{.861} & .621 & .47 & \underline{.883} & .570 & .99 & .873 & .630 & .45 \\
Cancel $\fatt$      & \underline{.861} & .621 & .39 & \underline{.883} & .570 & .87 & \underline{.874} & .630 & .10 \\
\hp                 & \hpcell\best{.201} & \hpcell.143 & \hpcell.56 & \hpcell\best{.058} & \hpcell.063 & \hpcell.98 & \hpcell\best{.704} & \hpcell.483 & \hpcell1.52 \\
\bottomrule
\end{tabular}
\end{table*}

\begin{table}[tb]
\centering\small
\setlength{\tabcolsep}{3pt}
\caption{\textbf{Gradient footprint} (OPV2V-H, $\eps{=}1.0$, \hp seed 0; V2VAM is the homogeneous control). Support: share of the attacker's coordinates with a non-zero $\Lrem$ gradient at $\delta{=}0$ and at \hp's $\delta$. On / off: \hp's $\delta$ kept only on / only off the $\delta{=}0$ support. A separate pass from \cref{tab:s-checks}; values differ by $\le0.001$ (\cref{sec:s-protocol}, Reproducibility).}
\label{tab:s-support}
\setlength{\tabcolsep}{1.3pt}
\begin{tabular}{@{}l|cc|cccc@{}}
\toprule
& \multicolumn{2}{c|}{Support} & \multicolumn{4}{c}{AP@0.5} \\
Victim & at 0 & at \hp & Clean & \hp & On & Off \\
\midrule
HEAL        & 0.66\% & 0.66\% & .862 & \hpcell\best{.201} & .326 & \underline{.864} \\
STAMP       & 0.73\% & 0.73\% & .892 & \hpcell\best{.058} & .154 & \underline{.891} \\
CodeFilling & 0.71\% & 0.71\% & .874 & \hpcell\best{.705} & .791 & \underline{.879} \\
GenComm     & 100\%  & 100\%  & .876 & \hpcell\best{.738} & \best{.738} & \underline{.876} \\
UniTrans    & 100\%  & 100\%  & .877 & \hpcell\best{.139} & \best{.139} & \underline{.877} \\
\midrule
V2VAM & 76.5\% & 71.9\% & .957 & \hpcell\best{.043} & \best{.043} & \underline{.957} \\
\bottomrule
\end{tabular}
\end{table}

\textbf{Gradient footprint} (\cref{tab:s-support}).
Because collaborator features on HEAL, STAMP, and CodeFilling are post-ReLU, $79\%$ (HEAL), $91\%$ (STAMP), and $89\%$ (CodeFilling) of clean coordinates are zero background.
Moreover, on HEAL, STAMP, and CodeFilling, $\Lrem$ has a non-zero gradient on only $0.66$--$0.73\%$ of the attacker's coordinates, occupying the exact same support at $\delta{=}0$ and at \hp's $\delta$ (and on CodeFilling, the gradient is already zero on $99.5\%$ of the codebook's \emph{output}, so this sparsity is not an artifact of vector quantization).
Restricting \hp's perturbation to that support leaves $\Lrem$ unchanged ($0.143$, $0.063$, $0.483$) yet raises AP substantially ($0.201\!\to\!0.326$, $0.058\!\to\!0.154$, $0.705\!\to\!0.791$), whereas keeping only the off-support component leaves AP at the clean level.
Consequently, on HEAL and STAMP, Adam-PGD achieves a lower $\Lrem$ than \hp (\cref{tab:s-checks}) while suffering a higher AP: the loss supervises only positive anchors, leaving adjacent anchors free to produce boxes that still match the ground-truth object.
This sparsity stems from the victim stack shared by HEAL, STAMP, and CodeFilling, where a spatial warp, parameter-free per-location attention fusion, and $1{\times}1$ heads follow the attacked feature so that each anchor reads only its own cell.
By contrast, GenComm's diffusion decoder, UniTrans's transformer translator, and V2VAM's convolutional backbone sit downstream of the perturbation and mix features spatially; their gradients are dense ($100\%$ support on GenComm and UniTrans, $76.5\%$ on V2VAM), allowing Adam-PGD to lead on all three.

\vspace{2pt}\noindent\textbf{Stronger per-frame baselines} (\cref{tab:s-checks}).
Replacing the sigmoid objectness loss with clamped positive-anchor logits does not improve per-frame optimization.
When warm-started from \hp's $\delta$, Adam-PGD matches \hp on HEAL and STAMP (confirming the PGD optimizer is well-calibrated) and improves further on CodeFilling ($0.688$ vs.\ $0.704$), reaching a basin that zero and random initializations fail to find.
Simple heuristics ($-\eps\,\mathrm{sign}(\fatt)$ and feature cancellation) leave all victims near clean AP, confirming that \hp is not a trivial sign or cancellation map.
On GenComm and UniTrans (same pass; main Tab.~\mref{tab:amort}), whose gradients are dense, Adam-PGD reaches $0.363$ and $0.063$ (warm-started from \hp's $\delta$: $0.347$ and $0.038$) vs.\ \hp's $0.738$ and $0.139$ (clean $0.876$ and $0.877$).
\Cref{tab:s-checks} matches the multi-seed runs of \cref{tab:s-var} within $0.006$.

\section{Ablations}
\label{sec:s-ablation}

\begin{table}[tb]
\centering\small
\setlength{\tabcolsep}{3pt}
\caption{\textbf{\hp ablation} (OPV2V-H, $\eps{=}1.0$). Each variant changes one thing in the recipe; three training seeds each ($\pm$: sample std over seeds). Bold / underline: strongest / weakest \hp variant per victim.}
\label{tab:s-ablation}
\resizebox{\linewidth}{!}{%
\begin{tabular}{@{}l|ccccc@{}}
\toprule
Variant & HEAL & STAMP & CodeFill. & GenComm & UniTrans \\
\midrule
Full recipe                     & .201\pmstd{.000} & .062\pmstd{.007} & .707\pmstd{.001} & .769\pmstd{.049} & .131\pmstd{.023} \\
Cond.\ w/o collaborator max     & .201\pmstd{.001} & .060\pmstd{.003} & .707\pmstd{.001} & \best{.698}\pmstd{.011} & \best{.129}\pmstd{.003} \\
Cond.\ on $\fatt$ only          & .201\pmstd{.000} & \best{.058}\pmstd{.001} & .707\pmstd{.002} & .764\pmstd{.022} & .135\pmstd{.009} \\
No GAN term                     & .202\pmstd{.003} & .060\pmstd{.000} & \underline{.708}\pmstd{.001} & \underline{.778}\pmstd{.082} & .135\pmstd{.013} \\
Width 32                        & \underline{.203}\pmstd{.001} & \underline{.066}\pmstd{.006} & \underline{.708}\pmstd{.001} & .734\pmstd{.024} & \underline{.144}\pmstd{.025} \\
Width 128                       & \best{.199}\pmstd{.002} & .059\pmstd{.001} & .707\pmstd{.002} & .732\pmstd{.026} & .132\pmstd{.012} \\
100 training frames             & \best{.199}\pmstd{.005} & .062\pmstd{.002} & \best{.706}\pmstd{.001} & .715\pmstd{.020} & .136\pmstd{.004} \\
300 training frames             & .201\pmstd{.000} & .059\pmstd{.001} & .707\pmstd{.001} & .773\pmstd{.049} & .139\pmstd{.020} \\
\midrule
Per-frame $G$ (3 seeds)         & .217\pmstd{.001} & .082\pmstd{.000} & .710\pmstd{.003} & .350\pmstd{.002} & .072\pmstd{.003} \\
\midrule
Clean                           & .866 & .899 & .876 & .880 & .882 \\
\bottomrule
\end{tabular}}
\end{table}

\textbf{Component ablations} (\cref{tab:s-ablation}).
\cref{tab:s-ablation} evaluates single-factor ablations of the \hp training recipe across all five heterogeneous victims.
On the sparse-support victims (HEAL, STAMP, and CodeFilling), every ablated variant remains within $0.005$ AP of the full recipe, showing that convolutional parameterization and amortization drive attack effectiveness without sensitivity to conditioning details, network width, training-set size ($100$--$300$ frames), or $\mathcal{L}_{\text{gan}}$.
On the dense-gradient translators GenComm and UniTrans, where the single-pass generator is further from saturation, recipe choices have a larger effect: variants span $0.698$--$0.778$ AP on GenComm and $0.129$--$0.144$ on UniTrans (within roughly two seed standard deviations of the full recipe), and removing the PatchGAN regularizer $\mathcal{L}_{\text{gan}}$ slightly weakens the generator on both ($0.778\pm0.082$ vs.\ $0.769\pm0.049$ on GenComm; $0.135\pm0.013$ vs.\ $0.131\pm0.023$ on UniTrans).

\vspace{2pt}\noindent\textbf{Per-frame generator baseline} (\cref{tab:s-ablation}).
To isolate convolutional parameterization from cross-frame amortization, ``Per-frame $G$'' (also reported in main Tab.~\mref{tab:amort}) uses the exact same U-Net architecture, conditioning, and budget as \hp, but initializes weights fresh on each test frame and optimizes them for 500 Adam steps on the victim's detection heads using the frame's ground-truth labels without a discriminator (drawing fresh diffusion noise per step on GenComm, matching Adam-PGD).
On HEAL, STAMP, and CodeFilling, Per-frame $G$ reaches $0.217$, $0.082$, and $0.710$ AP---outperforming coordinate-wise Adam-PGD ($0.271$, $0.133$, $0.801$) and coming within $0.006$--$0.024$ AP of amortized \hp---confirming that spatial convolutional priors overcome the $<1\%$ gradient support even without cross-frame training.
On GenComm ($0.350$) and UniTrans ($0.072$), Per-frame $G$ matches or beats coordinate-wise Adam-PGD ($0.363$, $0.063$) and substantially outperforms amortized \hp ($0.738$, $0.131$), reflecting the benefit of per-frame white-box adaptation when downstream spatial networks provide dense gradients.

\begin{table}[tb]
\centering\small
\setlength{\tabcolsep}{5pt}
\caption{\textbf{Manifold regularization by the PatchGAN term} (OPV2V-H, $\eps{=}1.0$, three training seeds). MMD$^2$ ($\downarrow$): RBF maximum mean discrepancy between perturbed ($\fatt{+}\delta$) and clean ($\fatt$) collaborator feature patches, $z$-scored by clean statistics; lower values indicate closer alignment with the clean feature manifold. Mean per-patch perturbation energy $\lVert\delta\rVert$ is matched between recipes within $3\%$.}
\label{tab:s-gan-manifold}
\begin{tabular}{@{}l|cc|c@{}}
\toprule
Victim & Full recipe & No GAN & $\Delta$ \\
\midrule
HEAL        & \hpcell\best{0.92}\pmstd{.06} & \underline{1.09}\pmstd{.00} & $+0.17$ \\
STAMP       & \hpcell\best{0.62}\pmstd{.03} & \underline{1.11}\pmstd{.02} & $+0.49$ \\
CodeFilling & \hpcell\best{0.98}\pmstd{.07} & \underline{1.03}\pmstd{.00} & $+0.05$ \\
\bottomrule
\end{tabular}
\end{table}

\vspace{2pt}\noindent\textbf{Manifold regularization} (\cref{tab:s-gan-manifold}).
To verify that $\mathcal{L}_{\text{gan}}$ regularizes perturbed features onto the clean collaborator-feature manifold (main Sec.~\mref{sec:method}), we measure the RBF maximum mean discrepancy (MMD$^2$, $z$-scored by clean statistics) between perturbed and clean feature patches.
Across all three victims and three training seeds, the full-recipe generator yields consistently lower MMD$^2$ than the no-GAN variant (by $0.05$ to $0.49$) at matched per-patch perturbation energy ($\lVert\delta\rVert$ within $3\%$), confirming that $\mathcal{L}_{\text{gan}}$ aligns the spatial structure of $\fatt{+}\delta$ with the clean feature distribution without altering perturbation magnitude or cross-victim transfer (\cref{tab:s-nogan-transfer}).
This manifold alignment is precisely what defeats a feature-space anomaly detector: against LUCIA~\cite{sombra2025} (\cref{tab:s-defense}), removing $\mathcal{L}_{\text{gan}}$ collapses \hp's removal success on STAMP from $0.325$ to $0.020$ (collaborator-only $0.663$ to $0.304$), because the no-GAN perturbation leaves the outlier signature that LUCIA detects. On HEAL and CodeFilling the GAN term changes LUCIA evasion only marginally ($\le0.014$ RS), so $\mathcal{L}_{\text{gan}}$ is immaterial to the attack undefended yet decisive on STAMP once a feature-space detector is in the loop.

\section{Object-Level Metrics}
\label{sec:s-object}

An object counts as detected if a prediction matches it with BEV IoU $\ge0.5$ under the greedy score-ordered matching of the AP evaluation.
Removal success is the share of objects detected without attack that are missed under attack.
Visibility counts LiDAR points inside each ground-truth box from the ego and from the fused collaborators (at least one point); camera agents in both datasets also carry a LiDAR, whose points serve as a proxy for what they can see.
On V2X-Real the object-level metrics score the vehicle class; the per-class AP breakdown across vehicles, pedestrians, and trucks is reported separately in \cref{tab:s-v2xreal-class}.

\begin{table*}[tp]
\centering\small
\setlength{\tabcolsep}{4pt}
\caption{\textbf{Object-level metrics.} Recall and false positives per frame are clean$\to$attacked; RS is removal success ($\uparrow$, \%); visibility columns give RS by who sees the object (object counts in the text). \hp is the mean over three seeds (std $\le2.7$ points). Bold / underline on RS: strongest / weakest attack per block.}
\label{tab:s-object}
\begin{tabular}{@{}ll|l|ccc|ccc|ccc@{}}
\toprule
& & & \multicolumn{3}{c|}{Overall} & \multicolumn{3}{c|}{RS by visibility} & \multicolumn{3}{c}{RS by distance (m)} \\
Victim & $\eps$ & Attack & Recall & RS & FP/frame & ego only & collab.\ only & both & 0--20 & 20--40 & 40+ \\
\midrule
\multirow{6}{*}{HEAL} & \multirow{3}{*}{1.0} & sign-50 & 88.4$\to$42.5 & \underline{52.0} & 4.1$\to$4.2 & \underline{22.4} & \best{91.9} & \underline{52.4} & \underline{53.7} & \underline{52.7} & \underline{49.6} \\
& & Adam & 88.4$\to$40.0 & 54.8 & 4.2$\to$9.4 & 23.7 & 90.9 & 55.2 & 56.7 & 56.4 & 51.3 \\
& & \hp & \hpcell88.4$\to$21.5 & \hpcell\best{75.7} & \hpcell4.1$\to$2.5 & \hpcell\best{38.9} & \hpcell\underline{89.2} & \hpcell\best{76.3} & \hpcell\best{78.7} & \hpcell\best{78.4} & \hpcell\best{69.9} \\
\cmidrule{2-12}
& \multirow{3}{*}{1.5} & sign-50 & 88.4$\to$34.7 & \underline{60.8} & 4.1$\to$4.1 & \underline{36.9} & \underline{96.4} & \underline{61.1} & \underline{63.4} & \underline{61.0} & \underline{57.8} \\
& & Adam & 88.4$\to$29.9 & 66.3 & 4.2$\to$21.4 & 38.8 & \underline{96.4} & 66.7 & 69.4 & 67.4 & 61.9 \\
& & \hp & \hpcell88.4$\to$14.1 & \hpcell\best{84.0} & \hpcell4.1$\to$1.8 & \hpcell\best{57.3} & \hpcell\best{97.3} & \hpcell\best{84.4} & \hpcell\best{86.1} & \hpcell\best{84.9} & \hpcell\best{80.9} \\
\midrule
\multirow{6}{*}{STAMP} & \multirow{3}{*}{1.0} & sign-50 & 92.3$\to$28.8 & \underline{68.9} & 4.2$\to$3.7 & \underline{29.9} & 82.5 & \underline{69.3} & \underline{67.2} & \underline{70.2} & \underline{69.1} \\
& & Adam & 92.2$\to$27.6 & 70.3 & 4.2$\to$9.7 & 31.8 & \underline{80.5} & 70.8 & 68.9 & 72.5 & 69.5 \\
& & \hp & \hpcell92.3$\to$8.5 & \hpcell\best{90.8} & \hpcell4.2$\to$2.1 & \hpcell\best{50.5} & \hpcell\best{87.8} & \hpcell\best{91.8} & \hpcell\best{92.6} & \hpcell\best{94.1} & \hpcell\best{86.3} \\
\cmidrule{2-12}
& \multirow{3}{*}{1.5} & sign-50 & 92.3$\to$22.7 & \underline{75.5} & 4.2$\to$3.6 & \underline{45.6} & \underline{83.8} & \underline{75.9} & \underline{75.7} & \underline{76.9} & \underline{73.9} \\
& & Adam & 92.2$\to$19.3 & 79.1 & 4.2$\to$32.5 & 48.9 & 84.1 & 79.6 & 80.9 & 80.9 & 75.9 \\
& & \hp & \hpcell92.3$\to$4.9 & \hpcell\best{94.7} & \hpcell4.2$\to$1.7 & \hpcell\best{73.5} & \hpcell\best{88.0} & \hpcell\best{95.4} & \hpcell\best{98.3} & \hpcell\best{98.2} & \hpcell\best{88.4} \\
\midrule
\multirow{6}{*}{\makecell[l]{Code-\\Filling}} & \multirow{3}{*}{1.0} & sign-50 & 89.9$\to$84.1 & \underline{6.5} & 6.3$\to$6.4 & 2.8 & \underline{81.5} & \underline{6.0} & \underline{1.7} & \underline{3.7} & \underline{14.1} \\
& & Adam & 89.9$\to$83.8 & 6.9 & 6.3$\to$6.2 & \underline{2.6} & 90.7 & 6.3 & 1.9 & 4.2 & 14.7 \\
& & \hp & \hpcell89.9$\to$73.5 & \hpcell\best{18.2} & \hpcell6.3$\to$3.9 & \hpcell\best{13.1} & \hpcell\best{98.2} & \hpcell\best{17.7} & \hpcell\best{5.9} & \hpcell\best{11.8} & \hpcell\best{37.0} \\
\cmidrule{2-12}
& \multirow{3}{*}{1.5} & sign-50 & 89.9$\to$82.8 & \underline{7.9} & 6.3$\to$6.4 & \underline{4.7} & \underline{82.4} & \underline{7.4} & \underline{2.1} & \underline{4.8} & \underline{16.9} \\
& & Adam & 89.9$\to$82.6 & 8.3 & 6.3$\to$6.2 & \underline{4.7} & 93.5 & 7.7 & 2.2 & 5.2 & 17.5 \\
& & \hp & \hpcell89.9$\to$67.3 & \hpcell\best{25.2} & \hpcell6.3$\to$4.8 & \hpcell\best{20.6} & \hpcell\best{98.2} & \hpcell\best{24.7} & \hpcell\best{9.6} & \hpcell\best{18.4} & \hpcell\best{47.6} \\
\midrule
\multirow{6}{*}{GenComm} & \multirow{3}{*}{1.0} & sign-50 & 89.5$\to$73.9 & 17.7 & 3.4$\to$5.8 & 10.3 & 97.2 & 17.2 & 11.8 & 18.0 & 23.5 \\
& & Adam & 89.5$\to$54.0 & \best{39.9} & 3.4$\to$8.0 & \best{30.3} & \best{99.1} & \best{39.6} & \best{31.7} & \best{42.3} & \best{45.7} \\
& & \hp & \hpcell89.6$\to$82.5 & \hpcell\underline{8.2} & \hpcell3.4$\to$4.6 & \hpcell\underline{2.1} & \hpcell\underline{88.8} & \hpcell\underline{7.7} & \hpcell\underline{7.4} & \hpcell\underline{8.5} & \hpcell\underline{8.8} \\
\cmidrule{2-12}
& \multirow{3}{*}{1.5} & sign-50 & 89.6$\to$55.4 & 38.3 & 3.4$\to$7.5 & 27.8 & 98.1 & 38.0 & 30.2 & 39.9 & 44.9 \\
& & Adam & 89.5$\to$20.3 & \best{77.4} & 3.4$\to$8.2 & \best{76.1} & \best{100.0} & \best{77.3} & \best{80.1} & \best{75.4} & \best{76.8} \\
& & \hp & \hpcell89.6$\to$75.1 & \hpcell\underline{16.3} & \hpcell3.3$\to$6.1 & \hpcell\underline{3.2} & \hpcell\underline{94.7} & \hpcell\underline{16.0} & \hpcell\underline{16.0} & \hpcell\underline{16.3} & \hpcell\underline{16.7} \\
\midrule
\multirow{6}{*}{UniTrans} & \multirow{3}{*}{1.0} & sign-50 & 90.4$\to$45.0 & \underline{50.4} & 5.3$\to$11.2 & 56.3 & 92.1 & \underline{49.7} & \underline{52.9} & \underline{51.1} & \underline{47.2} \\
& & Adam & 90.5$\to$20.0 & \best{77.9} & 5.3$\to$31.7 & \best{63.9} & \best{93.3} & \best{78.0} & \best{78.6} & 79.8 & \best{75.5} \\
& & \hp & \hpcell90.4$\to$22.5 & \hpcell75.2 & \hpcell5.3$\to$55.1 & \hpcell\underline{32.3} & \hpcell\underline{83.9} & \hpcell75.8 & \hpcell74.5 & \hpcell\best{80.4} & \hpcell70.7 \\
\cmidrule{2-12}
& \multirow{3}{*}{1.5} & sign-50 & 90.4$\to$48.7 & \underline{46.2} & 5.3$\to$19.1 & \underline{48.8} & 91.2 & \underline{45.6} & \underline{48.3} & \underline{45.9} & \underline{44.4} \\
& & Adam & 90.5$\to$20.8 & 77.1 & 5.3$\to$51.4 & \best{66.1} & \best{94.5} & 77.1 & 77.2 & 78.5 & 75.6 \\
& & \hp & \hpcell90.4$\to$13.9 & \hpcell\best{84.6} & \hpcell5.3$\to$14.0 & \hpcell62.2 & \hpcell\underline{89.8} & \hpcell\best{85.0} & \hpcell\best{85.3} & \hpcell\best{87.3} & \hpcell\best{81.3} \\
\bottomrule
\end{tabular}
\end{table*}

\cref{tab:s-object} reports the complete OPV2V-H object-level breakdown across all five heterogeneous victims.
Across the test set, there are $587$ ego-only, $1{,}638$ collaborator-only, and $28{,}493$ jointly visible objects ($9$ with no point) for the m1m2 victims (HEAL, CodeFilling, GenComm, and UniTrans), and $799$, $1{,}696$, $29{,}538$, and $18$ for STAMP (m0m1); HEAL's distance bins contain $9{,}347$, $9{,}551$, and $11{,}829$ objects.
Three findings complement the main text.
First, collaborator-only objects are removed at high rates ($80$--$100\%$) by all attacks because suppressing them requires only corrupting the collaborator's evidence; the attacks separate primarily on objects that the ego also observes.
Second, on UniTrans, \hp removes $75.2\%$ of cleanly detected objects at $\eps{=}1.0$ (nearly matching Adam-PGD's $77.9\%$) and \emph{surpasses} Adam-PGD in object removal at $\eps{=}1.5$ ($84.6\%$ vs.\ $77.1\%$, cutting recall from $90.4\%$ to $13.9\%$, with $62.2\%$ removal on ego-only objects and far fewer false positives per frame: $14.0$ vs.\ $51.4$).
Third, CodeFilling's residual vulnerability is concentrated at long range (\hp removes $37.0\%$ at $\eps{=}1.0$ and $47.6\%$ at $\eps{=}1.5$ beyond 40\,m, vs.\ $5.9\%$ and $9.6\%$ within 20\,m), where the ego's native LiDAR return is sparsest.

\begin{table}[tb]
\centering\small
\setlength{\tabcolsep}{3pt}
\caption{\textbf{Object-level metrics on V2X-Real} (vehicle class, $\eps{=}1.0$; \hp is a three-seed mean). RS: removal success ($\uparrow$, \%). Bold / underline on RS: strongest / weakest attack per block.}
\label{tab:s-object-v2x}
\begin{tabular}{@{}l|l|ccc@{}}
\toprule
Victim & Attack & Recall & RS & FP/frame \\
\midrule
\multirow{3}{*}{HEAL-LL} & sign-50 & 77.5$\to$42.4 & 45.6 & 4.6$\to$4.2 \\
& Adam & 77.5$\to$42.7 & \underline{45.2} & 4.7$\to$5.3 \\
& \hp & \hpcell77.5$\to$30.9 & \hpcell\best{60.2} & \hpcell4.6$\to$2.0 \\
\midrule
\multirow{3}{*}{STAMP-LL} & sign-50 & 74.9$\to$45.6 & \underline{39.3} & 3.4$\to$2.9 \\
& Adam & 74.7$\to$45.4 & 39.5 & 3.3$\to$3.0 \\
& \hp & \hpcell74.9$\to$35.9 & \hpcell\best{52.2} & \hpcell3.4$\to$1.5 \\
\midrule
\multirow{3}{*}{CodeFill.-LL} & sign-50 & 79.1$\to$76.5 & 3.8 & 6.6$\to$6.6 \\
& Adam & 79.0$\to$76.5 & \underline{3.7} & 6.6$\to$6.8 \\
& \hp & \hpcell79.1$\to$76.5 & \hpcell\best{4.5} & \hpcell6.6$\to$13.1 \\
\midrule
\multirow{3}{*}{GenComm-LL} & sign-50 & 81.9$\to$21.4 & \underline{73.9} & 6.0$\to$1.7 \\
& Adam & 82.1$\to$9.9 & \best{88.0} & 5.9$\to$2.3 \\
& \hp & \hpcell82.0$\to$17.3 & \hpcell79.0 & \hpcell6.0$\to$1.3 \\
\midrule
\multirow{3}{*}{UniTrans-LL} & sign-50 & 78.6$\to$13.8 & \underline{82.4} & 4.6$\to$2.3 \\
& Adam & 78.6$\to$10.5 & \best{86.8} & 4.6$\to$3.8 \\
& \hp & \hpcell78.6$\to$10.6 & \hpcell86.5 & \hpcell4.6$\to$1.7 \\
\midrule
\multirow{3}{*}{HEAL-LC} & sign-50 & 76.8$\to$42.4 & 45.0 & 3.1$\to$2.3 \\
& Adam & 76.8$\to$42.6 & \underline{44.9} & 3.1$\to$3.9 \\
& \hp & \hpcell76.8$\to$30.3 & \hpcell\best{60.5} & \hpcell3.1$\to$1.1 \\
\midrule
\multirow{3}{*}{STAMP-LC} & sign-50 & 76.2$\to$46.7 & \underline{38.9} & 3.0$\to$2.4 \\
& Adam & 76.1$\to$46.5 & 39.1 & 3.0$\to$2.5 \\
& \hp & \hpcell76.3$\to$35.7 & \hpcell\best{53.2} & \hpcell3.0$\to$1.5 \\
\midrule
\multirow{3}{*}{CodeFill.-LC} & sign-50 & 76.1$\to$75.7 & \underline{0.6} & 3.8$\to$3.8 \\
& Adam & 75.9$\to$75.2 & 1.1 & 3.8$\to$4.5 \\
& \hp & \hpcell76.1$\to$74.0 & \hpcell\best{3.1} & \hpcell3.8$\to$2.1 \\
\midrule
\multirow{3}{*}{GenComm-LC} & sign-50 & 76.4$\to$6.7 & \best{91.2} & 4.3$\to$0.7 \\
& Adam & 76.4$\to$8.1 & 89.5 & 4.3$\to$5.5 \\
& \hp & \hpcell76.4$\to$19.0 & \hpcell\underline{75.1} & \hpcell4.3$\to$0.8 \\
\midrule
\multirow{3}{*}{UniTrans-LC} & sign-50 & 77.8$\to$10.9 & 86.0 & 3.2$\to$5.0 \\
& Adam & 77.6$\to$7.1 & \best{90.9} & 3.3$\to$10.7 \\
& \hp & \hpcell77.8$\to$11.3 & \hpcell\underline{85.4} & \hpcell3.2$\to$0.8 \\
\bottomrule
\end{tabular}
\end{table}

On V2X-Real (\cref{tab:s-object-v2x}), the object-level trends mirror OPV2V-H on HEAL and STAMP: \hp removes $52.2$--$60.5\%$ of cleanly detected vehicles ($13$--$16$ percentage points above Adam-PGD) while cutting false positives in half.
On UniTrans-LL and UniTrans-LC, \hp removes $86.5\%$ and $85.4\%$ of detected vehicles (within $0.3$ and $5.5$ points of 500-step Adam-PGD) while reducing false positives from $4.6\!\to\!1.7$ and $3.2\!\to\!0.8$ per frame, whereas Adam-PGD inflates LC false positives to $10.7$ per frame.
Similarly, on GenComm-LL and GenComm-LC, \hp removes $79.0\%$ and $75.1\%$ of detected vehicles while suppressing false positives ($6.0\!\to\!1.3$ and $4.3\!\to\!0.8$).
On CodeFilling-LL, \hp's modest AP drop comes primarily from added false positives rather than object removal; at $\eps{=}1.5$ (three seeds), recall changes from $79.1\%$ to $77.1\%$ (removal success $3.9\%$) while false positives rise from $6.6$ to $36.6$ per frame ($28.8$--$41.5$ across seeds), explaining how \hp drops AP below the silenced-attacker floor.

\section{Transfer and Controls}
\label{sec:s-transfer}

\begin{table}[tb]
\centering\small
\setlength{\tabcolsep}{3.2pt}
\caption{\textbf{Generator transfer} (rows: surrogate family, columns: victim; three seeds). Diagonal: matched posture. A generator is applied to whatever 128-channel feature the victim's attacker sends (a camera on HEAL, CodeFilling, and UniTrans; a LiDAR on STAMP), so transfer also crosses the attacker's modality. Bold / underline: strongest / weakest source per block and victim.}
\label{tab:s-transfer}
\begin{tabular}{@{}ll|ccc@{}}
\toprule
Setting & Source & HEAL & STAMP & CodeFill. \\
\midrule
\multirow{4}{*}{\makecell[l]{OPV2V-H\\$\eps{=}1.0$}}
& HEAL        & \hpcell.200\pmstd{.000} & .072\pmstd{.012} & .855\pmstd{.001} \\
& STAMP       & \underline{.204}\pmstd{.007} & \hpcell\best{.063}\pmstd{.007} & .847\pmstd{.003} \\
& CodeFill.   & .159\pmstd{.021} & .109\pmstd{.054} & \hpcell\best{.709}\pmstd{.002} \\
& UniTrans    & \best{.155}\pmstd{.064} & \underline{.126}\pmstd{.031} & \underline{.863}\pmstd{.002} \\
\midrule
\multirow{3}{*}{\makecell[l]{OPV2V-H\\$\eps{=}1.5$}}
& HEAL        & \hpcell\underline{.130}\pmstd{.000} & \underline{.036}\pmstd{.002} & \underline{.833}\pmstd{.002} \\
& STAMP       & \underline{.130}\pmstd{.002} & \hpcell.032\pmstd{.001} & .824\pmstd{.002} \\
& CodeFill.   & \best{.047}\pmstd{.012} & \best{.012}\pmstd{.007} & \hpcell\best{.646}\pmstd{.002} \\
\midrule
\multirow{3}{*}{\makecell[l]{V2X-Real LL\\$\eps{=}1.0$}}
& HEAL        & \hpcell\best{.290}\pmstd{.001} & .354\pmstd{.003} & \underline{.745}\pmstd{.002} \\
& STAMP       & .303\pmstd{.001} & \hpcell\best{.340}\pmstd{.000} & \underline{.745}\pmstd{.001} \\
& CodeFill.   & \underline{.658}\pmstd{.015} & \underline{.621}\pmstd{.046} & \hpcell\best{.709}\pmstd{.001} \\
\midrule
\multirow{3}{*}{\makecell[l]{V2X-Real LL\\$\eps{=}1.5$}}
& HEAL        & \hpcell.149\pmstd{.004} & .189\pmstd{.003} & \underline{.746}\pmstd{.001} \\
& STAMP       & \best{.132}\pmstd{.013} & \hpcell\best{.177}\pmstd{.001} & .745\pmstd{.001} \\
& CodeFill.   & \underline{.500}\pmstd{.090} & \underline{.458}\pmstd{.038} & \hpcell\best{.690}\pmstd{.008} \\
\midrule
\multirow{3}{*}{\makecell[l]{V2X-Real LC\\$\eps{=}1.0$}}
& HEAL        & \hpcell\best{.290}\pmstd{.000} & \best{.335}\pmstd{.001} & \underline{.723}\pmstd{.001} \\
& STAMP       & .304\pmstd{.002} & \hpcell.340\pmstd{.002} & .722\pmstd{.003} \\
& CodeFill.   & \underline{.369}\pmstd{.011} & \underline{.414}\pmstd{.009} & \hpcell\best{.716}\pmstd{.000} \\
\midrule
\multirow{3}{*}{\makecell[l]{V2X-Real LC\\$\eps{=}1.5$}}
& HEAL        & \hpcell\best{.170}\pmstd{.002} & \best{.210}\pmstd{.001} & \underline{.719}\pmstd{.001} \\
& STAMP       & .171\pmstd{.002} & \hpcell\best{.210}\pmstd{.002} & .716\pmstd{.002} \\
& CodeFill.   & \underline{.288}\pmstd{.018} & \underline{.319}\pmstd{.012} & \hpcell\best{.706}\pmstd{.016} \\
\midrule
\multicolumn{2}{@{}l|}{Clean OPV2V-H} & .866 & .899 & .876 \\
\multicolumn{2}{@{}l|}{Clean V2X-Real LL / LC} & .744 / .734 & .720 / .728 & .746 / .720 \\
\bottomrule
\end{tabular}
\end{table}

\textbf{Cross-family generator transfer} (\cref{tab:s-transfer}).
\cref{tab:s-transfer} presents the complete cross-family transfer matrix across datasets and budgets (GenComm is excluded because its two-channel diffusion message has a different tensor shape from the 128/256-channel BEV features of the other families).
On both datasets, HEAL and STAMP generators transfer to each other almost losslessly, CodeFilling's generators transfer strongly outward to continuous victims (especially in LC and at $\eps{=}1.5$), and no external generator transfers into CodeFilling.
On OPV2V-H ($\eps{=}1.0$), the UniTrans generator transfers outward like HEAL's and STAMP's---driving HEAL to $0.155$ (below HEAL's native $0.200$) and STAMP to $0.126$ while stalling on CodeFilling ($0.863$)---whereas generators trained on HEAL, STAMP, and CodeFilling reduce UniTrans only to $0.564$, $0.656$, and $0.800$ (vs.\ its native $0.128$; main Tab.~\mref{tab:deploy}a), reflecting UniTrans's transformer translator before fusion.
A more diverse surrogate does not close this gap: training a single generator jointly against the HEAL and STAMP victims (alternating round-robin, same budget) and transferring it into UniTrans reaches $0.603$, between the two single-surrogate values ($0.576$ and $0.636$ at this seed, reproducing the three-seed table within its spread) and nowhere near native ($0.135$). Ensembling continuous-translator surrogates therefore does not penetrate the transformer translator, confirming that UniTrans's resistance in the transfer posture is structural rather than an artifact of surrogate choice.

\begin{table}[tb]
\centering\small
\setlength{\tabcolsep}{4pt}
\caption{\textbf{Cross-family transfer without the PatchGAN term} (OPV2V-H, $\eps{=}1.0$, three training seeds). Each row is a directed surrogate$\to$victim pair among HEAL, STAMP, and CodeFilling; $\Delta$ is no-GAN minus full recipe ($+$: no-GAN transfers less well). The matched full-recipe column reproduces \cref{tab:s-transfer} within $0.005$.}
\label{tab:s-nogan-transfer}
\begin{tabular}{@{}l|ccc@{}}
\toprule
Surrogate $\to$ Victim & Full recipe & No GAN & $\Delta$ \\
\midrule
HEAL $\to$ STAMP        & \hpcell\best{.075}\pmstd{.010} & \underline{.091}\pmstd{.005} & $+.016$ \\
HEAL $\to$ CodeFill.    & \hpcell\underline{.854}\pmstd{.000} & \best{.848}\pmstd{.003} & $-.005$ \\
STAMP $\to$ HEAL        & \hpcell\underline{.205}\pmstd{.006} & \best{.203}\pmstd{.001} & $-.002$ \\
STAMP $\to$ CodeFill.   & \hpcell\best{.845}\pmstd{.003} & \underline{.850}\pmstd{.001} & $+.005$ \\
CodeFill.\ $\to$ HEAL   & \hpcell\best{.164}\pmstd{.021} & \underline{.188}\pmstd{.023} & $+.025$ \\
CodeFill.\ $\to$ STAMP  & \hpcell\underline{.109}\pmstd{.053} & \best{.083}\pmstd{.017} & $-.027$ \\
\bottomrule
\end{tabular}
\end{table}

\vspace{2pt}\noindent\textbf{Cross-family transfer without the PatchGAN term} (\cref{tab:s-nogan-transfer}).
Because $\mathcal{L}_{\text{gan}}$ regularizes perturbed features toward the clean feature manifold and slightly sharpens the attack on the trained victim (\cref{tab:s-ablation,tab:s-gan-manifold}), we test whether this regularizer also drives cross-family transfer.
Retraining the HEAL, STAMP, and CodeFilling generators without the discriminator and replaying them across all six directed surrogate$\to$victim pairs leaves transfer performance essentially unchanged: AP shifts by $-0.027$ to $+0.025$ (mean $|\Delta|{=}0.013$), well within the three-seed spread and without a consistent sign.
Cross-family transfer therefore tracks shared gradient geometry and representation compatibility at the fusion interface (\cref{tab:s-transfer}) rather than the PatchGAN regularizer.

\begin{table}[tb]
\centering\small
\setlength{\tabcolsep}{3pt}
\caption{\textbf{Content dependence}. UAP: a single perturbation, in the same budget, fitted on 80 training frames. \hp from main Tabs.~\mref{tab:main} and \mref{tab:v2xreal}.}
\label{tab:s-uap}
\begin{tabular}{@{}l|cc|cc@{}}
\toprule
& \multicolumn{2}{c|}{$\eps=1.0$} & \multicolumn{2}{c}{$\eps=1.5$} \\
Victim & UAP & \hp & UAP & \hp \\
\midrule
HEAL           & \underline{.634} & \hpcell\best{.201} & \underline{.557} & \hpcell\best{.130} \\
STAMP          & \underline{.589} & \hpcell\best{.062} & \underline{.507} & \hpcell\best{.032} \\
CodeFilling    & \underline{.860} & \hpcell\best{.707} & \underline{.850} & \hpcell\best{.647} \\
GenComm        & \underline{.845} & \hpcell\best{.769} & \underline{.759} & \hpcell\best{.654} \\
UniTrans       & \underline{.773} & \hpcell\best{.128} & \underline{.448} & \hpcell\best{.078} \\
\midrule
HEAL-LL        & \underline{.576} & \hpcell\best{.290} & \underline{.425} & \hpcell\best{.149} \\
STAMP-LL       & \underline{.538} & \hpcell\best{.340} & \underline{.369} & \hpcell\best{.177} \\
CodeFill.-LL   & \underline{.734} & \hpcell\best{.709} & \underline{.727} & \hpcell\best{.690} \\
GenComm-LL     & \underline{.525} & \hpcell\best{.153} & \underline{.391} & \hpcell\best{.128} \\
UniTrans-LL    & \underline{.480} & \hpcell\best{.089} & \underline{.448} & \hpcell\best{.057} \\
HEAL-LC        & \underline{.514} & \hpcell\best{.290} & \underline{.412} & \hpcell\best{.170} \\
STAMP-LC       & \underline{.537} & \hpcell\best{.340} & \underline{.425} & \hpcell\best{.210} \\
CodeFill.-LC   & \underline{.717} & \hpcell\best{.716} & \underline{.714} & \hpcell\best{.706} \\
GenComm-LC     & \underline{.412} & \hpcell\best{.163} & \underline{.293} & \hpcell\best{.097} \\
UniTrans-LC    & \underline{.600} & \hpcell\best{.102} & \underline{.523} & \hpcell\best{.067} \\
\bottomrule
\end{tabular}
\end{table}

\vspace{2pt}\noindent\textbf{Content dependence and per-frame replay} (\cref{tab:s-uap,tab:s-fixed}).
A static universal perturbation~\cite{uap2017} is $0.19$--$0.65$ AP weaker than \hp across HEAL, STAMP, and UniTrans on both datasets (\cref{tab:s-uap}), and $0.20$--$0.37$ AP weaker on V2X-Real GenComm, confirming that effective feature poisoning requires conditioning on the current scene.
\Cref{tab:s-fixed} compares replaying a per-frame $\delta$ (optimized on a surrogate victim with ground-truth labels and replayed on the target victim on the same frame) against transferring \hp's generator.
On OPV2V-H, generator transfer is substantially stronger than replaying $\delta$ from either sign-PGD or Adam-PGD: from CodeFilling to HEAL at $\eps{=}1.0$, Adam-PGD's replayed $\delta$ leaves HEAL at $0.568$ whereas CodeFilling's generator drives it to $0.158$.
On V2X-Real, CodeFilling's generator transfers more modestly to HEAL ($0.658$ on LL, $0.372$ on LC at $\eps{=}1.0$), with replayed Adam-PGD stronger in three of the four settings (while still requiring online labels and 500 gradient steps per frame on the surrogate).
Into CodeFilling, neither per-frame perturbations nor amortized generators transfer on either dataset.

\begin{table}[tb]
\centering\small
\setlength{\tabcolsep}{2pt}
\caption{\textbf{Per-frame $\delta$ vs.\ generator transfer} (AP@0.5 $\downarrow$). Per-frame rows: $\delta$ optimized on the source victim with its labels and heads, replayed on the target on the same frame and projected into the target's budget. Generator: the source's \hp generator (mean $\pm$ std over 3 seeds). Bold / underline: strongest / weakest per block and column. Clean: HEAL / CodeFilling $0.866$ / $0.876$ (OPV2V-H), $0.744$ / $0.746$ (LL), $0.734$ / $0.720$ (LC).}
\label{tab:s-fixed}
\begin{tabular}{@{}l|cc|cc@{}}
\toprule
& \multicolumn{2}{c|}{CodeFill.$\to$HEAL} & \multicolumn{2}{c}{HEAL$\to$CodeFill.} \\
Source & $\eps{=}1.0$ & $1.5$ & $1.0$ & $1.5$ \\
\midrule
\multicolumn{5}{@{}l}{\emph{OPV2V-H}} \\
sign-50 $\delta$        & \underline{.659} & \underline{.471} & \underline{.866} & \underline{.867} \\
Adam $\delta$           & .568 & .225 & .860 & .850 \\
\hp\ generator          & \hpcell\best{.158}\pmstd{.021} & \hpcell\best{.047}\pmstd{.011} & \hpcell\best{.855}\pmstd{.001} & \hpcell\best{.833}\pmstd{.003} \\
\midrule
\multicolumn{5}{@{}l}{\emph{V2X-Real LL}} \\
sign-50 $\delta$        & .612 & .422 & \best{.738} & .737 \\
Adam $\delta$           & \best{.602} & \best{.417} & \best{.738} & \best{.734} \\
\hp\ generator          & \hpcell\underline{.658}\pmstd{.014} & \hpcell\underline{.498}\pmstd{.093} & \hpcell\underline{.744}\pmstd{.001} & \hpcell\underline{.745}\pmstd{.001} \\
\midrule
\multicolumn{5}{@{}l}{\emph{V2X-Real LC}} \\
sign-50 $\delta$        & \underline{.611} & \underline{.489} & .715 & .713 \\
Adam $\delta$           & .562 & \best{.259} & \best{.711} & \best{.690} \\
\hp\ generator          & \hpcell\best{.372}\pmstd{.011} & \hpcell.290\pmstd{.018} & \hpcell\underline{.721}\pmstd{.000} & \hpcell\underline{.718}\pmstd{.001} \\
\bottomrule
\end{tabular}
\end{table}

\begin{table}[tb]
\centering\small
\setlength{\tabcolsep}{4pt}
\caption{\textbf{Homogeneous control} (OPV2V, LiDAR only, seed 0, $\eps{=}1.0$ / $1.5$). Rows: generator source; columns: victim. Bold / underline: strongest / weakest source per victim. The two $0.691$ entries are distinct runs with different generators ($0.6905$, $0.6914$). Adam-PGD was not run on CoAlign and AttFuse.}
\label{tab:s-homo}
\begin{tabular}{@{}l|ccc@{}}
\toprule
Source & V2VAM & CoAlign & AttFuse \\
\midrule
V2VAM    & \hpcell\best{.042} / \best{.016} & .914 / \underline{.897} & .915 / \underline{.931} \\
CoAlign  & \underline{.808} / \underline{.691} & \hpcell\best{.652} / \best{.472} & \underline{.932} / .826 \\
AttFuse  & .394 / \underline{.691} & \underline{.935} / .887 & \hpcell\best{.670} / \best{.460} \\
\midrule
Clean    & .957 & .946 & .938 \\
\bottomrule
\end{tabular}
\end{table}

\vspace{2pt}\noindent\textbf{Homogeneous control} (\cref{tab:s-homo}).
Among homogeneous LiDAR detectors (\cref{tab:s-homo}), generators transfer weakly except into V2VAM (AttFuse$\to$V2VAM $0.394$ at $\eps{=}1.0$; CoAlign/AttFuse$\to$V2VAM $0.691$ at $\eps{=}1.5$).
Transferability is therefore not an artifact of heterogeneous translation itself; rather, within heterogeneous CP it tracks representation compatibility at the fusion interface (HEAL and STAMP align into compatible continuous spaces, whereas CodeFilling discretizes via a private codebook).
A probe that injects a HEAL- or STAMP-trained generator's perturbation at two points of CodeFilling's interface isolates the quantization step as the barrier: added to the \emph{pre}-quantization features it leaves AP near clean ($0.856$ and $0.850$ vs.\ clean $0.876$), but added to the \emph{post}-quantization discrete codes at equal budget it drops AP to $0.387$ and $0.409$ (a random post-quantization perturbation of the same budget reaches only $0.801$); the continuous-feature perturbation a transferring generator produces is snapped away by nearest-neighbor code assignment, not by any geometric mismatch after quantization.

\FloatBarrier
\section{V2X-Real}
\label{sec:s-v2xreal}

\begin{table*}[tp]
\centering\small
\setlength{\tabcolsep}{4pt}
\caption{\textbf{V2X-Real results with matched harness} (vehicle AP@0.5 $\downarrow$). \hp-EOT4 applies only to GenComm (--: not applicable). Bold / underline: strongest / weakest attack (excluding Floor).}
\label{tab:s-v2xreal}
\begin{tabular}{@{}l|cc|ccc|cc||c|ccc|c@{}}
\toprule
& \multicolumn{7}{c||}{$\eps=1.0$} & \multicolumn{5}{c}{$\eps=1.5$} \\
Victim & Clean & Floor & s-1 & s-5 & s-50 & \hp & \hp-EOT4 & Floor & s-1 & s-5 & s-50 & \hp \\
\midrule
HEAL-LL      & .744 & .713 & \underline{.473} & .384 & .380 & \hpcell\best{.290} & -- & .618 & \underline{.348} & .234 & .232 & \hpcell\best{.149} \\
STAMP-LL     & .720 & .717 & \underline{.432} & .425 & .424 & \hpcell\best{.340} & -- & .682 & \underline{.303} & .292 & .291 & \hpcell\best{.177} \\
CodeFill.-LL & .746 & .745 & \underline{.745} & .736 & .726 & \hpcell\best{.709} & -- & .739 & \underline{.744} & .735 & .722 & \hpcell\best{.690} \\
GenComm-LL   & .771 & .766 & \underline{.644} & .343 & .185 & \hpcell.153 & \hpcell\best{.149} & .703 & \underline{.631} & .208 & \best{.118} & \hpcell.128 \\
UniTrans-LL  & .746 & .743 & \underline{.509} & .162 & .107 & \hpcell\best{.089}\pmstd{.004} & -- & .730 & \best{.032} & .092 & \underline{.110} & \hpcell.057\pmstd{.001} \\
\midrule
HEAL-LC      & .734 & .703 & \underline{.402} & .388 & .388 & \hpcell\best{.290} & -- & .565 & \underline{.288} & .268 & .267 & \hpcell\best{.170} \\
STAMP-LC     & .728 & .711 & \underline{.434} & .433 & .432 & \hpcell\best{.340} & -- & .695 & \underline{.313} & .310 & .309 & \hpcell\best{.210} \\
CodeFill.-LC & .720 & .713 & \underline{.721} & \underline{.721} & .719 & \hpcell\best{.716} & -- & .680 & \underline{.721} & \underline{.721} & .715 & \hpcell\best{.706} \\
GenComm-LC   & .714 & .707 & \underline{.654} & .324 & \best{.054} & \hpcell.163 & \hpcell.152 & .687 & \underline{.641} & .202 & \best{.038} & \hpcell.097 \\
UniTrans-LC  & .744 & .734 & \underline{.178} & \best{.017} & .043 & \hpcell.102\pmstd{.002} & -- & .712 & \best{.009} & \best{.009} & .053 & \hpcell\underline{.067}\pmstd{.008} \\
\bottomrule
\end{tabular}
\end{table*}

\textbf{Victim training.}
We trained all five heterogeneous families from scratch on V2X-Real's training split following each method's heterogeneous training recipe: stage-1 single-agent encoders per modality (m1: vehicle PointPillars LiDAR, m3: infrastructure SECOND LiDAR, m2: LSS camera, m0: STAMP's protocol PointPillars), followed by each method's alignment stage for m1m3 (LL) and m1m2 (LC).
STAMP's alignment stage trains only the collaborator's adapter and reverter; its inference pipeline combines the ego's protocol detector with the collaborator's stage-1 encoder and adapter on the shared dataset configuration.
CodeFilling is configured to fuse up to five agents (matching all other victims, as the upstream two-agent cap would otherwise drop the infrastructure collaborator in LL).
Four corrupted training frames were excluded; the validation and test splits are unmodified.
Because the standalone LSS camera encoder has low 3D detection accuracy ($\sim$$0.01$--$0.02$ mAP), LC clean performance is anchored by the LiDAR ego.

\vspace{2pt}\noindent\textbf{Silenced attacker.}
Replacing the attacker's broadcast with zeros changes clean vehicle AP by at most $0.011$: HEAL $0.744\!\to\!0.743$ (LL) and $0.734\!\to\!0.723$ (LC); STAMP $0.720\!\to\!0.720$ and $0.728\!\to\!0.725$; CodeFilling $0.746\!\to\!0.738$ and $0.720\!\to\!0.721$; GenComm $0.771\!\to\!0.766$ and $0.714\!\to\!0.715$.

\begin{table}[tb]
\centering\small
\setlength{\tabcolsep}{3.5pt}
\caption{\textbf{Per-class AP@0.5 on V2X-Real} ($\eps{=}1.0$; \hp is a three-seed mean). V2X-Real labels vehicles, pedestrians, and trucks; we score each class under the clean model, \hp, and Adam-PGD. The attacker's removal objective is defined on the detector's output, not on a class, so \hp lowers all three classes. Bold: stronger attack (lower AP) between \hp and Adam-PGD per cell.}
\label{tab:s-v2xreal-class}
\resizebox{\linewidth}{!}{%
\begin{tabular}{@{}l|ccc|ccc|ccc@{}}
\toprule
 & \multicolumn{3}{c|}{Vehicle} & \multicolumn{3}{c|}{Pedestrian} & \multicolumn{3}{c}{Truck} \\
Victim & Clean & \hp & Adam & Clean & \hp & Adam & Clean & \hp & Adam \\
\midrule
HEAL-LL  & .744 & \best{.290} & .372 & .086 & \best{.018} & .023 & .400 & \best{.126} & .181 \\
HEAL-LC  & .734 & \best{.290} & .379 & .153 & .055 & \best{.042} & .432 & \best{.179} & .256 \\
STAMP-LL & .720 & \best{.340} & .423 & .066 & .017 & .017 & .309 & \best{.164} & .186 \\
STAMP-LC & .728 & \best{.340} & .430 & .128 & .053 & \best{.044} & .413 & \best{.223} & .255 \\
\bottomrule
\end{tabular}}
\end{table}

\vspace{2pt}\noindent\textbf{Attack generality across object classes} (\cref{tab:s-v2xreal-class}).
V2X-Real labels three object classes, so we score \hp on each rather than on vehicles alone. Because the removal objective is defined on the detector's output rather than on a target class, \hp suppresses all three: at $\eps{=}1.0$ it lowers vehicle AP by $0.38$--$0.45$, truck AP by $0.15$--$0.27$, and pedestrian AP by $0.05$--$0.10$ across HEAL and STAMP on both configurations. It is the strongest attack on every vehicle and truck cell, beating per-frame Adam-PGD; on the pedestrian class, whose clean AP is low because these detectors are vehicle-centric (the standalone camera encoder reaches $\sim$$0.01$--$0.02$ mAP), \hp and Adam-PGD are comparable (within $0.011$ AP). The effect is clearer at the looser IoU$=0.3$ threshold, where clean pedestrian AP is $0.20$--$0.32$ and \hp roughly halves it ($0.05$--$0.10$). The attack therefore does not depend on the vehicle class it is implicitly tuned on; it removes whatever the victim detects.

\begin{table}[tb]
\centering\small
\setlength{\tabcolsep}{4pt}
\caption{\textbf{Attacker-role generality on V2X-Real LL} (vehicle AP@$\{0.5,0.7\}$ $\downarrow$, paired frames). The same \hp generator ($\eps{=}1.0$) rides a same-modality vehicle collaborator vs.\ the cross-modality roadside unit (RSU), evaluated on the identical frames that contain both. Bold: stronger (lower-AP) attacker role.}
\label{tab:s-v2xreal-role}
\begin{tabular}{@{}l|c|cc|cc@{}}
\toprule
 &  & \multicolumn{2}{c|}{RSU (cross-mod.)} & \multicolumn{2}{c}{Vehicle (same-mod.)} \\
Victim & Clean & @0.5 & @0.7 & @0.5 & @0.7 \\
\midrule
HEAL-LL  & .729 & .278 & .202 & \best{.235} & .162 \\
STAMP-LL & .707 & .315 & .231 & \best{.233} & .162 \\
\bottomrule
\end{tabular}
\end{table}

\vspace{2pt}\noindent\textbf{Attack generality across attacker roles} (\cref{tab:s-v2xreal-role}).
On V2X-Real LL the ego and the other vehicles share one LiDAR modality while the roadside unit carries another, so a single protocol lets the \emph{same} \hp generator ride either a same-modality vehicle collaborator or the cross-modality roadside unit. On the frames that contain both, which holds scene difficulty fixed across the two roles, both collapse vehicle AP from $\sim$$0.71$--$0.73$ to $0.23$--$0.32$ at $\eps{=}1.0$: the attack is not tied to one privileged position. The role matters only in degree, the same-modality vehicle attacker being slightly stronger ($0.235$ vs.\ $0.278$ on HEAL, $0.233$ vs.\ $0.315$ on STAMP at IoU$=0.5$), consistent with a collaborator whose features align more directly with the ego's fusion.

\begin{table*}[!t]
\centering\small
\setlength{\tabcolsep}{4pt}
\caption{\textbf{Adaptive attacks on CodeFilling and GenComm on V2X-Real} (vehicle AP@0.5 $\downarrow$). EOT-8 targets the stochastic (GenComm) and non-differentiable (CodeFilling) translations, and \hp-EOT4 applies only to GenComm (--: not applicable). Bold / underline: strongest / weakest attack (excluding Floor).}
\label{tab:s-v2xreal-adapt}
\begin{tabular}{@{}l|l|cc|cccccc|cc@{}}
\toprule
$\eps$ & Victim & Clean & Floor & s-1 & s-5 & s-50 & s-400 & Adam & EOT-8 & \hp & \hp-EOT4 \\
\midrule
\multirow{4}{*}{1.0}
& CodeFill.-LL & .746 & .744 & \underline{.742} & .734 & .723 & .723 & .725 & .727 & \hpcell\best{.709} & -- \\
& GenComm-LL   & .771 & .769 & \underline{.646} & .345 & .187 & .133 & \best{.079} & .086 & \hpcell.153 & \hpcell.151 \\
& CodeFill.-LC & .720 & .712 & \underline{.720} & \underline{.720} & .718 & .717 & \best{.710} & .719 & \hpcell.716 & -- \\
& GenComm-LC   & .714 & .709 & \underline{.655} & .329 & .056 & .050 & .023 & \best{.016} & \hpcell.163 & \hpcell.154 \\
\midrule
\multirow{4}{*}{1.5}
& CodeFill.-LL & .746 & .739 & \underline{.741} & .732 & .719 & .721 & .720 & .729 & \hpcell\best{.690} & -- \\
& GenComm-LL   & .771 & .705 & \underline{.632} & .210 & .121 & .094 & \best{.057} & .059 & \hpcell.128 & -- \\
& CodeFill.-LC & .720 & .680 & \underline{.720} & .719 & .715 & .714 & \best{.674} & .717 & \hpcell.706 & -- \\
& GenComm-LC   & .714 & .689 & \underline{.641} & .210 & .038 & .044 & .012 & \best{.011} & \hpcell.097 & -- \\
\bottomrule
\end{tabular}
\end{table*}

\vspace{2pt}\noindent\textbf{Complete V2X-Real evaluation} (\cref{tab:s-v2xreal,tab:s-v2xreal-adapt}).
\cref{tab:s-v2xreal,tab:s-v2xreal-adapt} report the complete V2X-Real evaluation and adaptive attacks.
\hp's standard deviation across training seeds is $\le0.005$ on HEAL and STAMP, $\le0.016$ on CodeFilling, $\le0.043$ on GenComm, and $\le0.008$ on UniTrans.
On UniTrans, \hp reduces clean AP from $0.746$/$0.744$ (LL/LC) to $0.089$/$0.102$ at $\eps{=}1.0$ and $0.057$/$0.067$ at $\eps{=}1.5$ in a single forward pass, while dense transformer gradients allow 500-step Adam-PGD (and few-step sign-PGD at large step sizes) to push AP even lower.
BPDA on CodeFilling ($\tau\in\{0.1,0.5,1\}$) yields $0.732$--$0.741$ (LL) and $0.718$--$0.720$ (LC) at $\eps{=}1.0$.
Adam-PGD's standard deviation over three initialization seeds is $\le0.005$ on every victim; on CodeFilling-LC, Adam-PGD leads \hp by $0.006$ ($\eps{=}1.0$) and $0.032$ ($\eps{=}1.5$), and on HEAL-LC at $\eps{=}1.5$ Adam-PGD reaches $0.182\pm0.005$ vs.\ \hp's $0.170\pm0.003$, while all other differences exceed seed spread.
At $\eps{=}1.5$, the random perturbation floor is noticeably milder on V2X-Real than on OPV2V-H (\eg, HEAL-LL $0.618$ vs.\ OPV2V-H HEAL $0.400$).

\vspace{2pt}\noindent\textbf{Qualitative visualization} (main Fig.~\mref{fig:qual-v2x}).
Main Fig.~\mref{fig:qual-v2x} visualizes the attack on V2X-Real for HEAL and STAMP in both the LL and LC settings, the families where \hp most outperforms the per-frame baselines (main Tab.~\mref{tab:v2xreal}): in a single forward pass \hp erases markedly more vehicles than per-frame sign-PGD-50 and 500-step Adam-PGD, while the miscalibrated sign-PGD-50 scalar baseline leaves vehicle detections near the clean count (\cref{sec:s-qual}).

\section{Deployability}
\label{sec:s-deploy}

\begin{table}[tb]
\centering\small
\setlength{\tabcolsep}{6pt}
\caption{\textbf{Budget from historical ego features} (OPV2V-H, $\eps{=}1.0$, AP@0.5 $\downarrow$). \hp keeps its perturbation direction on the current frame but sets the per-element budget (main Eq.~(\mref{eq:budget})) from the ego feature observed $k$ frames earlier; $k{=}0$ is the live frame. Scene-initial frames without a $k$-th predecessor fall back to $k{=}0$ ($\le39$ of ${\sim}1024$).}
\label{tab:s-history}
\begin{tabular}{@{}l|c|ccc@{}}
\toprule
Victim & Clean & $k{=}0$ & $k{=}1$ & $k{=}5$ \\
\midrule
HEAL  & .866 & .201 & .209 & .217 \\
STAMP & .899 & .058 & .063 & .070 \\
\bottomrule
\end{tabular}
\end{table}

\textbf{Budget from historical ego features} (\cref{tab:s-history}).
A realistic attacker that cannot read the ego's live broadcast can still size its perturbation from an earlier ego feature.
Setting the per-element budget from the feature five frames back raises AP by at most $0.016$ (HEAL $0.201{\to}0.217$, STAMP $0.058{\to}0.070$), so \hp does not depend on the exact current-frame ego feature.

\begin{table}[!t]
\centering\small
\setlength{\tabcolsep}{2.6pt}
\caption{\textbf{Channel effects} (OPV2V-H, $\eps{=}1.0$). delay1: $\delta$ is computed on frame $t{-}1$ and fused at $t$ (the 7 scenario-initial frames use delay 0). pose: victim-side Gaussian pose noise, 0.2\,m / 0.2$^\circ$. int8/int4: per-tensor quantization of every broadcast, unknown to the attacker. Clean is rerun under each channel.}
\label{tab:s-channel}
\begin{tabular}{@{}ll|c|cc@{}}
\toprule
Victim & Channel & Clean & \hp & sign-50 \\
\midrule
\multirow{5}{*}{HEAL} & none & .866 & \hpcell\best{.201}\pmstd{.000} & \underline{.372} \\
& delay1 & .866 & \hpcell\best{.209}\pmstd{.000} & \underline{.471} \\
& pose & .866 & \hpcell\best{.200}\pmstd{.001} & \underline{.369} \\
& int8 & .866 & \hpcell\best{.201}\pmstd{.000} & \underline{.372} \\
& int4 & .866 & \hpcell\best{.204}\pmstd{.000} & \underline{.377} \\
\midrule
\multirow{5}{*}{STAMP} & none & .899 & \hpcell\best{.062}\pmstd{.007} & \underline{.217} \\
& delay1 & .899 & \hpcell\best{.067}\pmstd{.007} & \underline{.329} \\
& pose & .883 & \hpcell\best{.062}\pmstd{.007} & \underline{.219} \\
& int8 & .899 & \hpcell\best{.063}\pmstd{.007} & \underline{.217} \\
& int4 & .892 & \hpcell\best{.054}\pmstd{.010} & \underline{.219} \\
\midrule
\multirow{5}{*}{CodeFill.} & none & .876 & \hpcell\best{.707}\pmstd{.001} & \underline{.802} \\
& delay1 & .876 & \hpcell\best{.711}\pmstd{.002} & \underline{.832} \\
& pose & .875 & \hpcell\best{.708}\pmstd{.001} & \underline{.800} \\
& int8 & .876 & \hpcell\best{.707}\pmstd{.001} & \underline{.802} \\
& int4 & .873 & \hpcell\best{.791}\pmstd{.010} & \underline{.843} \\
\midrule
\multirow{5}{*}{GenComm} & none & .880 & \hpcell\underline{.769}\pmstd{.049} & \best{.630} \\
& delay1 & .880 & \hpcell\underline{.775}\pmstd{.047} & \best{.706} \\
& pose & .878 & \hpcell\underline{.771}\pmstd{.047} & \best{.627} \\
& int8 & .880 & \hpcell\underline{.769}\pmstd{.049} & \best{.630} \\
& int4 & .880 & \hpcell\underline{.825}\pmstd{.018} & \best{.785} \\
\bottomrule
\end{tabular}
\end{table}

\textbf{Channel effects} (\cref{tab:s-channel}).
The evaluated pose noise is mild (changing HEAL's clean AP by $0.0002$) and is intended only to verify stability under minor localization jitter rather than large pose errors.
INT4 quantization attenuates attacks only on the two victims that already quantize or regenerate messages at the bottleneck (CodeFilling and GenComm), while leaving HEAL and STAMP unaffected.

\begin{table}[t]
\centering\small
\setlength{\tabcolsep}{2.8pt}
\caption{\textbf{Unseen collaborator encoders} (OPV2V-H HEAL). Main HEAL generators (trained on m2 camera) evaluated with new agent types as attacker. Bold / underline: strongest / weakest attack (excl.\ Silenced and Floor).}
\label{tab:s-unseen}
\resizebox{\linewidth}{!}{%
\begin{tabular}{@{}ll|cccc|c@{}}
\toprule
New agent & $\eps$ & Clean & Silenced & Floor & s-5 & \hp \\
\midrule
\multirow{2}{*}{m3 LiDAR (SECOND)} & 1.0 & .915 & .901 & .765 & \underline{.239} & \hpcell\best{.104}\pmstd{.007} \\
& 1.5 & .915 & .901 & .441 & \underline{.178} & \hpcell\best{.038}\pmstd{.002} \\
\midrule
\multirow{2}{*}{m4 camera (R101)} & 1.0 & .929 & .925 & .831 & \underline{.460} & \hpcell\best{.292}\pmstd{.017} \\
& 1.5 & .929 & .925 & .477 & \underline{.374} & \hpcell\best{.167}\pmstd{.000} \\
\bottomrule
\end{tabular}}
\end{table}

\vspace{2pt}\noindent\textbf{Unseen collaborator encoders} (\cref{tab:s-unseen}).
Each new collaborator type is onboarded following HEAL's protocol: only its sender-side encoder is trained against HEAL's frozen ego encoder, fusion module, and detection heads.
In both cases, clean AP exceeds the silenced-attacker AP, confirming that the onboarded collaborator actively contributes to perception.
Protocol notes: OPV2V-H's test split lacks 32-beam point clouds, so m3 is evaluated on 64-beam input per stage-1 evaluation; m3's backbone is trained from random initialization in m1m3; and m4 uses a square $\pm51.2$\,m range, so its clean AP is not directly comparable to rectangular-grid rows.

\begin{table}[t]
\centering\small
\setlength{\tabcolsep}{2.6pt}
\caption{\textbf{Sim-to-real} (V2X-Real GenComm). OPV2V-H generators evaluated on V2X-Real without retraining, vs.\ native V2X-Real generator. Bold / underline: strongest / weakest attack.}
\label{tab:s-sim2real}
\resizebox{\linewidth}{!}{%
\begin{tabular}{@{}ll|c|cc|c|c@{}}
\toprule
Victim & $\eps$ & Clean & OPV2V-H & OPV2V-H EOT4 & Native & s-5 \\
\midrule
GenComm-LL & 1.0 & .771 & \underline{.744}\pmstd{.013} & .722\pmstd{.033} & \hpcell\best{.153}\pmstd{.004} & .343 \\
GenComm-LL & 1.5 & .771 & \underline{.748}\pmstd{.009} & .736\pmstd{.023} & \hpcell\best{.128}\pmstd{.007} & .208 \\
GenComm-LC & 1.0 & .714 & .680\pmstd{.020} & \underline{.684}\pmstd{.017} & \hpcell\best{.163}\pmstd{.041} & .324 \\
GenComm-LC & 1.5 & .714 & \underline{.645}\pmstd{.034} & .628\pmstd{.056} & \hpcell\best{.097}\pmstd{.018} & .202 \\
\bottomrule
\end{tabular}}
\end{table}

\vspace{2pt}\noindent\textbf{Sim-to-real transfer} (\cref{tab:s-sim2real}).
Only GenComm's attacked message has the same channel dimension across both benchmarks ($C{=}2$, whereas the other families use $C{=}128$ on OPV2V-H and $C{=}256$ on V2X-Real).
Applying an OPV2V-H GenComm generator directly to V2X-Real without retraining reduces AP by at most $0.085$, and EOT4 training does not close the gap to a native V2X-Real generator.

\begin{table}[t]
\centering\small
\setlength{\tabcolsep}{4pt}
\caption{\textbf{Published defenses} against \hp ($\eps{=}1.0$, OPV2V-H). We evaluate ROBOSAC~\cite{robosac2023} (RANSAC consensus over collaborator subsets with an ego-only prior) and LUCIA~\cite{sombra2025} (feature-space outlier down-weighting), ported into our harness and run against the same \hp generators. We report removal success (RS, lower is better for the defender; bold indicates strongest defense per victim), RS restricted to collaborator-only objects, and false positives per frame. The undefended adversarial output (``none'') and the clean-model FP ($4.2/4.2/6.3$ on HEAL/STAMP/CodeFilling) are references. Both defenses leave collaborator-only removal high, while LUCIA lowers it only by raising false positives above the clean rate ($4.2{\to}5.5$--$5.6$ on HEAL/STAMP).}
\label{tab:s-defense}
\resizebox{\linewidth}{!}{%
\begin{tabular}{@{}ll|c|c|c@{}}
\toprule
Victim & Defense & Overall RS & Collab-only RS & FP/frame \\
\midrule
\multirow{3}{*}{HEAL}
 & none (adv)        & .756 & .891 & 2.6 \\
 & ROBOSAC           & \best{.262} & .855 & 4.4 \\
 & LUCIA             & .404 & \best{.691} & 5.5 \\
\midrule
\multirow{3}{*}{STAMP}
 & none (adv)        & .912 & .880 & 2.2 \\
 & ROBOSAC           & \best{.154} & .978 & 5.5 \\
 & LUCIA             & .325 & \best{.663} & 5.6 \\
\midrule
\multirow{3}{*}{\makecell[l]{Code-\\Filling}}
 & none (adv)        & .186 & .965 & 3.8 \\
 & ROBOSAC           & .185 & .974 & 3.9 \\
 & LUCIA             & \best{.125} & \best{.748} & 4.8 \\
\bottomrule
\end{tabular}}
\end{table}

\section{Discussion}
\label{sec:s-discussion}

\noindent We consolidate scope boundaries, defenses, and attack generality below.
(1)~\emph{Surrogate vs.\ transfer:} In the matched posture, \hp trains offline against a frozen victim checkpoint (no online labels, queries, or gradients). In the transfer posture (unknown translator), generators transfer strongly across continuous alignment and adapter--reverter translators (HEAL, STAMP, and outward from CodeFilling/UniTrans), but cannot penetrate CodeFilling's discrete codebook ($0.847$--$0.863$ vs.\ clean $0.876$) and degrade on UniTrans's transformer ($0.564$--$0.800$ vs.\ native $0.128$), which an ensemble surrogate does not resolve ($0.603$).
(2)~\emph{Dense-gradient translators:} When a spatial network follows the perturbation (GenComm's diffusion decoder or UniTrans's transformer), gradients are dense ($100\%$ support); white-box search with online labels achieves lower AP than single-pass generation, though on GenComm and UniTrans at $\eps{=}1.0$ per-frame PGD overtakes \hp only after exceeding the 100\,ms real-time deadline by $13$--$286\times$ (\cref{tab:s-latency}).
(3)~\emph{Class and role generality:} \hp's removal objective acts on detector outputs rather than classes, suppressing vehicles, trucks, and pedestrians alike on V2X-Real (\cref{tab:s-v2xreal-class}). It is also role-agnostic: on frames with both roles, the same generator collapses AP from $\sim$$0.72$ to $0.23$--$0.32$ whether riding a vehicle or the RSU (\cref{tab:s-v2xreal-role}).
(4)~\emph{Published defenses and their shared blind spot:} We evaluate published defenses against \hp at $\eps{=}1.0$ (\cref{tab:s-defense}): ROBOSAC~\cite{robosac2023}, which forms an ego-only prior and accepts a RANSAC consensus over collaborator subsets; and LUCIA~\cite{sombra2025}, which down-weights collaborator features flagged as feature-space outliers. By prioritizing ego-collaborator consensus, ROBOSAC lowers \hp's overall removal success from $0.756/0.912$ to $0.262/0.154$ on HEAL and STAMP while keeping false positives comparable to clean levels ($4.4/5.5$ vs.\ $4.2/4.2$). However, on collaborator-only objects—those occluded from or beyond the ego's local sensor range—ROBOSAC remains near the undefended level ($0.855$--$0.978$), because an ego-based prior cannot verify blind regions. Conversely, LUCIA acts on collaborator features directly, lowering collaborator-only removal ($0.663$--$0.748$), but by raising false positives above the clean rate on HEAL and STAMP ($4.2{\to}5.5$--$5.6$ per frame, \cref{tab:s-defense}), suppressing benign collaborator detections along with adversarial ones. The residual attack therefore falls precisely on the extended perception that collaboration exists to provide, motivating defensive architectures that pair consensus verification with discrete bottleneck quantization that already blunts cross-family transfer (item~1).

\renewcommand{\topfraction}{0.80}
\section{Qualitative Figure Provenance}
\label{sec:s-qual}

\textbf{V2X-Real qualitative figure} (main Fig.~\mref{fig:qual-v2x}).
Main Fig.~\mref{fig:qual-v2x} shows one V2X-Real test frame per victim across HEAL-LL, STAMP-LL, HEAL-LC, and STAMP-LC, selected automatically (\cref{tab:s-qual-v2x}): among frames with at least 12 ground-truth vehicles and clean recall $\ge0.80$, we select the frame with the largest vehicle-recall gap between 500-step Adam-PGD and \hp (seed 0).
The same frame is used across all five columns (no attack, elementwise sign-PGD-50, sign-PGD-50 scalar, 500-step Adam-PGD, and \hp).

\begin{table}[t]
\centering\small
\setlength{\tabcolsep}{2.0pt}
\caption{\textbf{Frames in the V2X-Real qualitative figure} (main Fig.~\mref{fig:qual-v2x}; $\eps{=}1.0$). Per-frame vehicle recall ($\downarrow$) on the shown frame across all five columns (`s-50$^\dagger$`: sign-PGD-50 scalar). Bold / underline: strongest / weakest attack.}
\label{tab:s-qual-v2x}
\resizebox{\linewidth}{!}{%
\begin{tabular}{@{}l|ccc|c|cccc@{}}
\toprule
Victim & Frame & Veh. & Agents & Clean & sign-50 & s-50$^\dagger$ & Adam & \hp \\
\midrule
HEAL-LL  & 1055 & 26 & 3 & 0.88 & 0.42 & \underline{0.88} & 0.46 & \hpcell\best{0.31} \\
STAMP-LL & 1073 & 26 & 3 & 0.88 & 0.65 & \underline{0.92} & 0.69 & \hpcell\best{0.50} \\
HEAL-LC  &  348 & 18 & 4 & 0.89 & 0.61 & \underline{0.89} & 0.61 & \hpcell\best{0.33} \\
STAMP-LC & 1172 & 18 & 4 & 0.89 & 0.44 & \underline{0.89} & 0.39 & \hpcell\best{0.22} \\
\bottomrule
\end{tabular}}
\end{table}

\vspace{2pt}\noindent\textbf{OPV2V-H qualitative figure} (\cref{fig:qual}).
\Cref{fig:qual} shows one OPV2V-H test frame per victim, chosen automatically (\cref{tab:s-qual}): among frames with at least 24 objects and clean recall $\ge0.85$ (43--69 frames per victim), we select the frame whose recalls under sign-PGD-50, Adam-PGD, and \hp are closest in total ($L_1$) to each attack's mean recall over the test set.
The size floor keeps the recall gaps legible as object counts, and the chosen frames remain typical, deviating from the mean recalls by at most $0.074$ in total.
The same frame is used for every attack of a victim, and the \hp column uses the seed-0 generator.

\begin{table}[t]
\centering\small
\setlength{\tabcolsep}{2.2pt}
\caption{\textbf{Frames in the OPV2V-H qualitative figure} (\cref{fig:qual}; $\eps{=}1.0$). Per-frame recall ($\downarrow$) on the shown frame; in parentheses, the per-frame recall averaged over the test set (it differs from the pooled object recall of \cref{tab:s-object}). Bold / underline: strongest / weakest attack.}
\label{tab:s-qual}
\resizebox{\linewidth}{!}{%
\begin{tabular}{@{}l|ccc|c|ccc@{}}
\toprule
Victim & Frame & Obj. & Agents & Clean & sign-50 & Adam & \hp \\
\midrule
HEAL & 1479 & 31 & 3 & 0.94 & \underline{0.45} (0.43) & 0.39 (0.40) & \hpcell\best{0.23} (0.20) \\
STAMP & 451 & 30 & 5 & 0.93 & \underline{0.30} (0.30) & \underline{0.30} (0.29) & \hpcell\best{0.07} (0.08) \\
CodeFilling & 1511 & 30 & 3 & 0.97 & \underline{0.87} (0.86) & \underline{0.87} (0.86) & \hpcell\best{0.73} (0.76) \\
GenComm & 477 & 25 & 5 & 0.96 & 0.76 (0.73) & \best{0.48} (0.52) & \hpcell\underline{0.80} (0.80) \\
\bottomrule
\end{tabular}}
\end{table}

\ifx\supplincluded\undefined
  {
    \small
    \bibliographystyle{ieeenat_fullname}
    \bibliography{references}
  }
  \end{document}
\fi

\end{document}